\documentclass{article}

\usepackage[final]{corl_2022} % Use this for the initial submission.
\usepackage{wrapfig}
\usepackage{graphicx}
\usepackage{amsmath}
\usepackage{amsfonts}
\usepackage{url}
\usepackage{algorithm}
\usepackage{algpseudocode}
\usepackage{caption}

\hypersetup{
    colorlinks=true,
    linkcolor=orange,
    filecolor=magenta,      
    urlcolor=orange,
    citecolor=orange,
}
\usepackage{pdfx}

\DeclareMathOperator{\E}{\mathbb{E}}

\DeclareMathOperator*{\argmin}{arg\,min}

\title{Few-Shot Preference Learning for Human-in-the-Loop RL}

\author{
  Joey Hejna\\
  Stanford University \\
  \texttt{jhejna@cs.stanford.edu} \\
   \And
   Dorsa Sadigh \\
   Stanford University \\
   \texttt{dorsa@cs.stanford.edu} \\
}

\begin{document}
\maketitle

%===============================================================================
\begin{abstract}
     While reinforcement learning (RL) has become a more popular approach for robotics, designing sufficiently informative reward functions for complex tasks has proven to be extremely difficult due their inability to capture human intent and policy exploitation. Preference based RL algorithms seek to overcome these challenges by directly learning reward functions from human feedback. Unfortunately, prior work either requires an unreasonable number of queries implausible for any human to answer or overly restricts the class of reward functions to guarantee the elicitation of the most informative queries, resulting in models that are insufficiently expressive for realistic robotics tasks. Contrary to most works that focus on query selection to \emph{minimize} the amount of data required for learning reward functions, we take an opposite approach: \emph{expanding} the pool of available data by viewing human-in-the-loop RL through the more flexible lens of multi-task learning. Motivated by the success of meta-learning, we pre-train preference models on prior task data and quickly adapt them for new tasks using only a handful of queries. Empirically, we reduce the amount of online feedback needed to train manipulation policies in Meta-World by 20$\times$, and demonstrate the effectiveness of our method on a real Franka Panda Robot. Moreover, this reduction in query-complexity allows us to train robot policies from actual human users. Videos of our results can be found at \url{https://sites.google.com/view/few-shot-preference-rl/home}.
\end{abstract}


%===============================================================================

\section{Introduction}
The success of deep reinforcement learning (RL) methods in game-playing and simulated domains \cite{mnih2013playing} has inspired recent work applying RL-based techniques to real-world robot control to middling success. Integral to the success of deep RL methods is the reward function, which describes the desired behavior of the learning agent. While training robots via trial and error holds great promise, designing suitable reward functions remains challenging. For example, consider teaching a robot to open a door. The simplest reward function would be sparse -- providing the robot with a positive reward only when the door has been opened. However, such sparse signals offer little learning signal, hampering exploration and enlarging sampling complexity. Conversely in designing a dense reward function, practitioners are tasked with summarizing multiple objectives like door angle or proximity to the handle into a single scalar. Such reward functions have proven to be difficult to design \cite{zhu2019ingredients} and can even cause agents to learn unintended behaviors. Hand-designed dense reward functions often do not directly parallel the goal-conditions humans want them to capture, causing RL agents to exploit them and potentially leading to hazardous policies that do not align with human intent \citep{hadfield2017inverse}. All of these problems are exacerbated in more realistic, multi-task scenarios with large state and action spaces \citep{pan2022the} where we might wish to teach agents how to complete a variety of tasks in their environment. A robot that can only open doors provides little utility in the real world. Given the effort required to design a single reward function, constructing reward functions for an entire family of tasks is impractical. \looseness=-1

Recent works attempt to circumvent the basic challenges of reward design by learning reward functions directly from human preferences. This paradigm has numerous advantages: learned reward functions are dense \citep{brys2015reinforcement, wu2021shaping}, easily aligned with human intent \citep{jeon2020reward}, and can be adapted \citep{xie2018few}. While demonstrations are often difficult to provide due to expensive data collection \citep{akgun2012keyframe} and large domain gaps \citep{losey2020controlling, smith2019avid}, human preferences can often be elicited solely through simple pairwise comparisons. However, given the large continuous state and action settings of robotics problems, learning a high-performance reward function from only a handful of noisy user generated binary labels seems hopeless \citep{Wright-Ma-2022}. Consequently, methods from active learning maximize feedback efficiency by attempting to ask the most informative queries with simplistic or linear reward models \citep{sadigh2017active, biyik2020active}. The constraints these methods place on the reward function class make them unable to scale to complex domains that necessitate expressive reward models \citep{christiano2017deep}. Moreover, such methods are not significantly more data efficient than random sampling in practice \citep{DBLP:journals/corr/abs-2107-02331, lowell2018practical}. On the other hand, recent works using general function approximators still require thousands to tens-of-thousands of artificially labeled queries to learn sufficiently accurate reward functions \citep{lee2021pebble, lee2021bpref, christiano2017deep}. This is far too onerous for real human labelers to provide, even in the single task setting. In order to train effective reward functions from actual humans, we need need a paradigm shift. Instead of optimizing for the most informative query, we take an orthogonal perspective that maximizes the amount of overall data by leveraging pre-training on realistic multi-task settings, and fine-tuning on a small and manageable amount of human queries online. \looseness=-1

In the multi-task setting, significantly more data is available from previously known tasks which can be used to accelerate reward function learning.  In fact, the shared structure of many real-world tasks has already been shown to accelerate policy learning \citep{yu2019meta}. The same structure can be exploited to learn complex reward functions for new tasks with only a handful of queries. This is largely because most tasks have rewards that are non-trivial compositions of other tasks. For example, data collected on opening windows and drawers could help us learn a reward function for door-opening with fewer human queries. Our key insight is to use multi-task data in order to meta-learn reward functions for preference based RL. Pre-training reward functions on a large dataset enables them to quickly adapt to new preferences with only a handful of queries.

Our core contributions are as follows. First we introduce a method for efficiently training RL policies from human-feedback using a meta-learned reward function. Second, we demonstrate its effectiveness across a number of standard robotics benchmarks, reducing query usage by a factor of 20 on robotic benchmarks in comparison to previous state-of-the-art methods. This increase in efficiency allows us to learn manipulation policies from real human feedback unlike prior work. Finally, we demonstrate the effectiveness of our method in the real-world using a Franka Panda robot.

\section{Related Work}
Our work builds on top of a number of prior works spanning RL, preference-learning, and meta-learning. Here we review the areas most relevant to our method.

\textbf{Reward Learning.} 
As hand-designed reward functions are difficult to tune, easily mis-specified \citep{hadfield2017inverse, turner2020avoiding}, and challenging to implement in the real world \citep{zhu2019ingredients, kormushev2010robot}, many recent works have leveraged human-collected data in order to learn reward functions. A large body of work focuses on using inverse RL, where a reward function is learned from approximately expert human collected demonstrations \citep{abbeel2004apprenticeship, ramachandran2007bayesian, ziebart2008maximum, brown2020better}. However, demonstration collection is often expensive \citep{khurshid2015data, akgun2012keyframe, dragan2012formalizing, losey2020controlling, losey2022learning} and collected demonstrations are sometimes not even aligned with true human preferences \citep{basu2017you, kwon2020humans, chen2020learning}. Alternative strategies for learning reward functions utilize physical corrections \citep{li2021learning}, natural language instructions \citep{co2018guiding}, human-provided scalar scores \citep{knox2009interactively, knox2008tamer} or partial \citep{myers2022learning} or complete \citep{brown2019extrapolating, biyik2019green} rankings . While physical corrections and language may be easier for the user, it is generally unclear how they translate to reward updates. Stronger signals are provided by scalar scores or multiple rankings, but they are harder for users to provide \textcolor{blue}{\citep{miller1956magical}}. We thus use pairwise comparisons as they are the simplest and generally refer to this approach as \textit{preference learning}. Many recent works have studied active preference-based learning from human feedback, however such approaches often make restrictive assumptions of the reward function, like linearity in predefined features \citep{sadigh2017active, pmlr-v164-wang22g, 6094735, biyik2020active,lepird2015bayesian}. These assumptions make such methods too inexpressive to scale to modern robot learning with complex objectives \textcolor{blue}{\citep{biyik2020active}}. While recent methods combining preference learning with deep RL make no assumptions on the structure of the reward function, they are far too feedback inefficient to be effectively used by humans \citep{christiano2017deep, lee2021pebble, park2022surf, liang2022reward, warnell2018deep}. Other works that use preferences with deep imitation learning \citep{brown2019deep} still require demonstrations. Most related to our work, PEBBLE \citep{lee2021pebble} combines the SAC off-policy RL algorithm \citep{haarnoja2018soft} with an ensemble of learned reward functions for sampling informative comparisons. Unfortunately, PEBBLE still requires an impractical number of queries to learn just a single task (25k for drawer opening). Distinct from prior work, we consider the more realistic multi-task setting that enables us to tap into a large amount of diverse data for for pre-training to increase query-efficiency. \looseness=-1

\textbf{Meta Learning}. Meta-learning methods \citep{finn2017model, nichol2018reptile} address the few-shot learning problem, where predictions on new tasks are made with a limited amount of data. Inspired by their success in supervised learning problems, we adopt the MAML algorithm \citep{finn2017model} for learning new reward functions based on a limited number of human queries. Though supervised meta-learning has been previously used to infer reward classifiers \citep{xie2018few} or in learning from heterogeneous demonstrators \citep{schrum2022personalized} to our knowledge it has not been applied to reward learning from preferences. Instead of adapting the reward function to new tasks, other related work in meta-RL directly adapts the policy network after a few exploratory episodes \citep{DBLP:conf/cogsci/WangKSLTMBKB17, agarwal2019learning, duan2016rl, rakelly2019efficient}. As the RL problem is much more difficult than supervised reward learning, policy adaptation approaches are likely to be less query efficient.

\section{Few-Shot Preference Learning for RL}
In this section we formally describe the problem of meta-learning for preference based RL, then detail how our algorithm leverages multi-task pre-training for online few-shot adaptation.
% the major components of our algorithm. 

\begin{figure}[t]
    \centering
    \includegraphics[width=\textwidth]{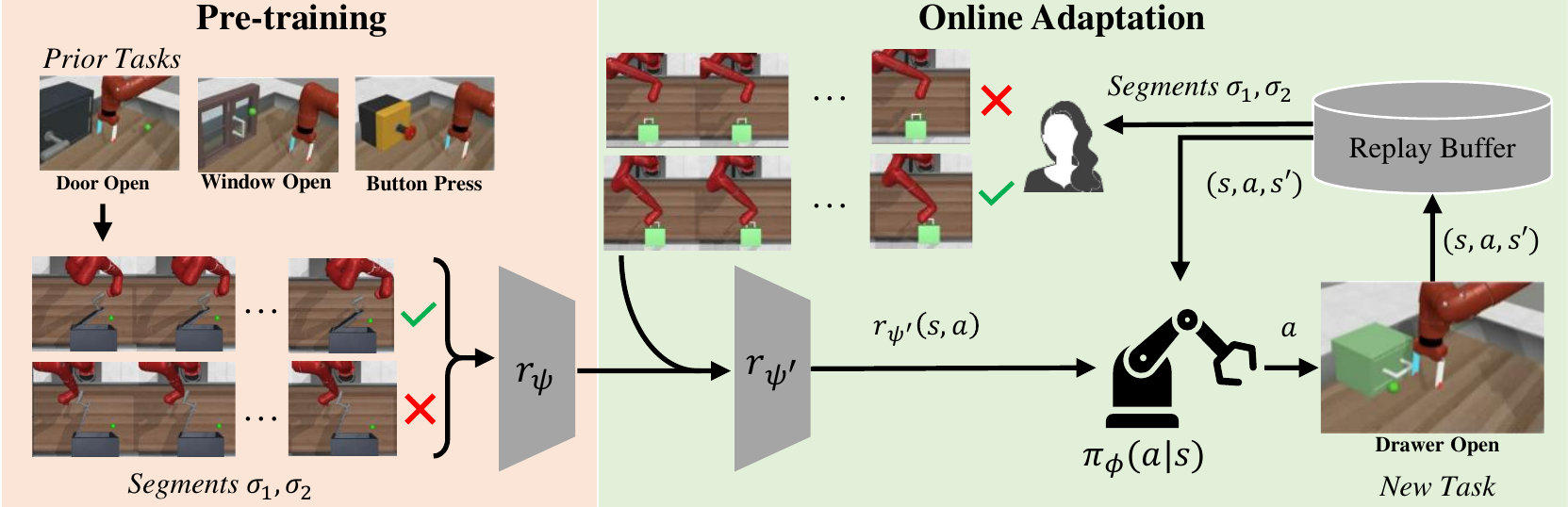}
    \vspace{-0.1in}
    \caption{An overview of our method. \textbf{Pre-training (left):} In the pre-training phase we generate trajectory segment comparisons using data from a family of previously learned tasks and use them to train a reward model. \textbf{Online-Adaptation (Right)}: After pre-training the reward model, we adapt it to new data from human feedback use it to train a policy for a new task in a closed loop manner. }
    \label{fig:method}
    \vspace{-0.2in}
\end{figure}

\textbf{Problem Setup.} In standard RL, an agent maximizes its cumulative expected reward in a Markov decision process (MDP). Unlike standard RL, we assume the reward function $r(s,a)$ to be unknown and instead must be estimated from human feedback. Distinct from prior works in preference based RL, we focus on the multi-task regime and thus additionally assume the existence of a distribution of tasks $p(\mathcal{T})$. Each task $\tau$ corresponds to a unique MDP where the state space $\mathcal{S}$, action space $\mathcal{A}$, and discount factor $\gamma$ are held constant, but the unknown ground-truth reward function $r(s,a)$ and sometimes transition function $\mathcal{P}$, vary. Thus, we write that $\tau_i = (\mathcal{P}_i, r_i) \sim p(\mathcal{T})$. 

Within this setting, we define the \textit{few-shot preference-based RL} problem. Given access to a dataset of $N$ previous tasks, $\{\tau_i\}_{i=1}^N$, the agents goal is to learn a policy $\pi_\text{new}(a|s)$ for a new task $\tau_\text{new} \sim p(\mathcal{T})$ from human feedback with as few user queries as possible. We make no explicit assumption on the form of prior data for each of the $N$ prior tasks, only that it contains sufficient information to learn an estimate of the reward $r_i$. After the pre-training phase, depicted in the left half of Figure \ref{fig:method}, we learn policies from online human feedback (right half of Figure \ref{fig:method}). This setting is a significant departure from past work in preference-based RL, as we do not assume that new tasks are learned in isolation. More realistically, there are multiple tasks that have been completed within the same state and action space. Next we explain the major components of our approach.

\textbf{Preference Learning.} In order to learn the policy $\pi_\text{new}(a|s)$ for a new task from human preferences, we choose to learn the new tasks' reward function $r_\text{new}(s,a)$. While alternative approaches might seek to directly adapt the policy $\pi \xrightarrow[]{} \pi_\text{new}$ using human feedback, such meta-RL style approaches often entail the difficult optimization challenges known to plague policy gradients and dynamic programming \citep{powell2016perspectives}. Instead, we directly model the reward using supervised learning techniques. We denote $\hat{r}_\psi(s,a)$ to be a learned estimate of an unknown ground-truth reward function $r(s,a)$, parameterized by $\psi$. As in \citet{wilson2012bayesian} we consider preferences over partial trajectory segments $\sigma = (s_t, a_t, s_{t+1}, a_{t+1}, ..., s_{t+k-1}, s_{t+k-1})$ of $k$ states and actions, as they provide more information than single states \citep{wilson2012bayesian, christiano2017deep}. We then define a preference predictor over segments using the Bradley-Terry model of paired comparisons \citep{bradley1952rank}:
\begin{equation*} \label{eq:prob}
    P[\sigma_1 \succ \sigma_2 ] = \frac{\exp \sum_t \hat{r}_\psi(s_t^{1}, a_t^{1})}{\exp \sum_t \hat{r}_\psi(s_t^{1}, a_t^{1}) + \exp \sum_t \hat{r}_\psi(s_t^{2}, a_t^{2})}
\end{equation*}
In the above, $\sigma_1 \succ \sigma_2$ indicates the event that segment 1 is preferred to segment 2, as shown in Figure \ref{fig:method}. For a given dataset $\mathcal{D}$ comprised of labeled queries $(\sigma_1, \sigma_2, y)$ where $y= \{1, 2\}$ corresponds to whether $\sigma_1$ or $\sigma_2$ is preferred, we optimize the following objective to learn $\hat{r}_\psi$.
\begin{equation} \label{eq:loss}
    \mathcal{L}_\text{pref}(\psi,  \mathcal{D}) = - \E_{(\sigma^1, \sigma^2, y) \sim \mathcal{D}} \left[ y(1) \log (P[\sigma_1 \succ \sigma_2 ]) + y(2)\log(1 - P[\sigma_1 \succ \sigma_2 ]) \right]
\end{equation}
In practice, this is just the standard binary cross-entropy objective where logits are determined by the sum of the learned reward function $\sigma$ over $k$ timesteps. Intuitively, this objective seeks to maximize the logits, and consequently predicted reward values, of the preferred segment in comparison to the unpreferred one. \looseness=-1

\begin{wrapfigure}{R}{0.5\textwidth}
\centering
\vspace{-0.35in}
\begin{minipage}[t]{.5\textwidth}
\begin{algorithm}[H]
\caption{Few-Shot Preference-based RL}
\label{alg}
\begin{algorithmic}[1]
\Require Teacher freq $K$, Queries per session $M$
\State $\psi \gets \argmin_\psi \sum_{i} \mathcal{L}\left(\psi - \alpha \nabla_\psi \mathcal{L}(\psi, \mathcal{D}_i), \mathcal{D}_i\right)$ 
\For{$t = 1, 2, 3, ...$}
\If{$t  \%  K == 0$}
    \For{$m = 1, 2, ... M$}
        \State $(\sigma_1, \sigma_2) \sim \text{Disagreement}$
        \State $y \gets$ user preference
        \State $\mathcal{D}_\text{new} \gets \mathcal{D}_\text{new} \cup (\sigma_1, \sigma_2,y)$
    \EndFor
    \State $\psi' \gets \psi$ Re-initialize reward model
    \For{each gradient step}
        \State $\psi' \gets \psi' - \alpha \nabla_{\psi'}   \mathcal{L}_\text{pref}(\psi', \mathcal{D}_\text{new})$
    \EndFor
\EndIf
\State Collect $s_{t+1}$ by taking $a_t \sim \pi(a_t|s_t)$
\State Store transition $\mathcal{B} \gets \mathcal{B} \cup (s_t, a_t, s_{t+1})$
\State Sample batch $\{(s_t, a_t, s_{t+1})\}_{j=1}^B \sim \mathcal{B}$
\State Assign rewards $r_t \gets r_{\psi'}(s_t,a_t))$
\State Optimize $\pi$ via SAC with \newline 
\hspace*{2em} $\{(s_t, a_t, s_{t+1}, r_{\psi'}(s_t,a_t))\}_{j=1}^B$
\EndFor
\end{algorithmic}
\end{algorithm}
\end{minipage}
\end{wrapfigure}

\textbf{Pre-training for Preference Learning.} To estimate the reward function of a new task $r_\text{new}$ in as few queries as possible, we want to pre-train a reward function $\hat{r}_\psi$ that can quickly adapt to new tasks with only a handful of comparisons $(\sigma_1, \sigma_2, y)$. Tapping into offline data can help exploit shared task structure and potential accelerate learning on new tasks. We propose extending the meta-learning framework to preference learning across different tasks. Our approach is agnostic to the choice of meta-learning algorithm, but we choose Model Agnostic Meta-Learning (MAML) \citep{finn2017model} for its simplicity. Concretely, MAML searches for parameters $\psi$ that attain high performance on a new task after only a few gradient steps by training on a set of previous tasks. In our setting, data for previous tasks can come from offline datasets, simulated policies, or actual humans. In conjunction with our preference loss from Equation \eqref{eq:loss}, we use the following pre-training update:
\begin{equation} \label{eq:maml}
    \psi \xleftarrow{} \psi - \beta \nabla_\psi \sum_{i = 1}^N \mathcal{L}_\text{pref} (\psi - \alpha \nabla_\psi \mathcal{L}_\text{pref}(\psi, \mathcal{D}_i), \mathcal{D}_i).
\end{equation}
Here $\alpha$ and $\beta$ are the inner and outer learning rates respectively. Each dataset $\mathcal{D}_i$ is comprised of known queries for each of the $N$ tasks $\tau_i \sim p(\mathcal{T})$. When we start training for a new task, we can quickly adapt the reward function using the new queries as $\psi' \xleftarrow{} \psi - \alpha \nabla_\psi \mathcal{L}_\text{pref}(\psi, \mathcal{D}_\text{new})$. As $\psi$ is explicitly optimized for performance on $\mathcal{L}_\text{pref}$ after only a handful of updates, we significantly reduce query complexity.\looseness=-1

Training $\hat{r}_\psi$ using Equation~\eqref{eq:maml} however, requires access to query datasets $\mathcal{D}_i$ for each task. While pre-training can be accomplished through several objectives, like reward regression, we use preference-based pre-training for consistency and its generality. Pairwise comparison data can be extracted from a wide variety of sources. If reward values are present in offline data, artificial labels $y$ for trajectory segments $\sigma_1$, $\sigma_2$ can easily be generated via the comparison $\sum_t r(s^1_t, a^1_t) > \sum_t r(s^2_t, a^2_t)$ as is common practice in prior works \citep{christiano2017deep, lee2021pebble}. If reward values for previous tasks are unknown but policies are, reward values can be recovered via inverse-RL, or comparisons can be derived from direct behavior comparison. For example, when generating queries for task $i$, behaviors from $\pi_i(a|s)$ would be preferred to behaviors generated from $\pi_{\ne i}(a|s)$. The left half of Figure \ref{fig:method} shows the process of extracting query data from offline data for pre-training, which corresponds to line 1 in Algorithm \ref{alg}. In our experiments, we use the artificial reward labeling scheme described first for consistency with prior work \citep{christiano2017deep}.

\textbf{Few-Shot Preference-based RL}. Our pre-trained preference function can then be used for few-shot preference based RL during an online adaptation phase, depicted in the right half of Figure \ref{fig:method}. We modify the standard Soft-Actor Critic RL algorithm \citep{haarnoja2018soft} to relabel transitions using our learned reward function before performing a standard actor-critic update (Algorithm \ref{alg} lines 17-18). Every $K$ steps, we ask a user to answer queries and provide feedback labels $y$ as shown in lines 5-7 of Algorithm \ref{alg}. Informative queries are selected using the disagreement of an ensemble of reward functions over the preference predictors. Specifically, comparisons that maximize $\text{std}(P[\sigma_1 \succ \sigma_2 ])$ are selected each time feedback is collected \citep{daniel2015active}. After new feedback is collected, we re-initialize the reward model $\hat{r}_\psi$ to its pre-trained weights. Subsequently, we re-adapt it using the updated dataset $\mathcal{D}_\text{new}$ for the new task as shown in Algorithm lines 9-11. 

To our knowledge, we are the first to leverage multi-task data for preference-based RL. The shift to the multi-task setting necessitates critical algorithmic changes in comparison with prior work. First, we pre-train the reward function from prior data instead of using other warm-start methods like unsupervised exploration used in PEBBLE. Second, we crucially reset the reward model for adaptation. Our setting provides a novel framework that leverages pre-training on a range of tasks for data-efficient adaptation on new tasks enabling human users to provide this data without making any structural assumptions on the reward function.

\begin{figure}[t]
    \centering
    \includegraphics[width=\textwidth]{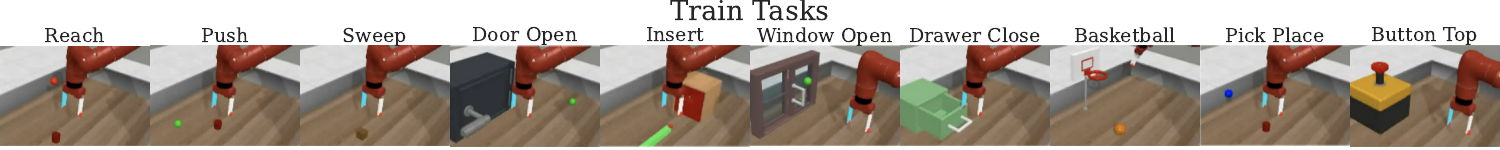}
    \includegraphics[width=\textwidth]{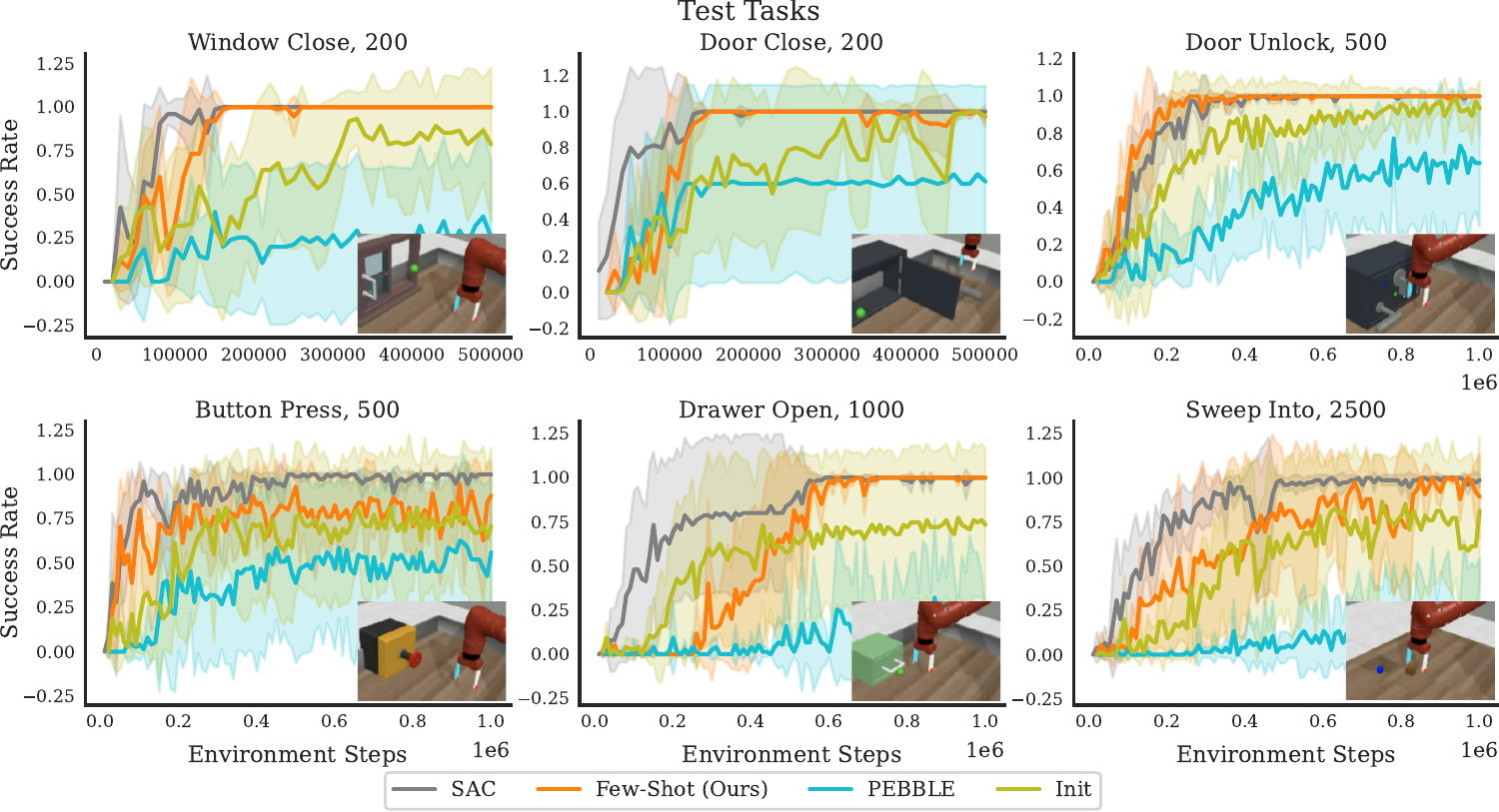}
    \vspace{-0.1in}
    \caption{Results on MetaWorld tasks. The title of each subplot indicates the task and number of artificial feedback queries used in training. Results for each method are shown across five seeds.}
    \label{fig:metaworld_results}
    \vspace{-0.15in}
\end{figure}

\section{Experiments}
In this section we seek to answer the following questions: First, does few-shot preference learning improve the query efficiency of preference-based RL? Second, is our method efficient enough to learn robot policies from real human feedback? Finally, can few-shot preference learning be used in the real world? Dataset, architecture, and hyperparameter details are available in the Appendix.

\subsection{How query-efficient is few-shot preference-based RL?}
\label{sec:simulation}
To test the query-efficiency of few-shot preference based RL for realistic robotic tasks, we adopt the Meta-World benchmark from \citet{yu2019meta}. Agent tasks include household activities like opening doors or closing windows, and standard manipulation problems like block pushing. Some Meta-World tasks are particularly difficult for human-in-the-loop learning as they are sequential: feedback on the second part of the task, like where an agent should move a block, can only be provided once the agent learns the first part of the task, like how to grasp a block. Additionally, different objects introducing different manipulation dynamics across tasks. To evaluate the raw-performance of our approach, we use the artificial queries induced by the task ground truth reward function. Previous works in preference based RL have required up to fifty-thousand artificial queries in order to solve some of the Meta-world tasks \citep{lee2021pebble}. Our approach generally achieves the same performance using 20$\times$ fewer queries. Our reward models are pre-trained using \textit{only 10 prior tasks} and evaluate query-efficiency on six previously unseen tasks. We compare our method, which we refer to as \textit{Few-Shot}, to three baselines: \looseness=-1
\begin{enumerate}
    \item \textbf{SAC}: The Soft-Actor Critic RL algorithm trained from ground truth rewards. This represents ``oracle'' performance. 
    \item \textbf{PEBBLE}: The PEBBLE algorithm from \citet{lee2021pebble}, which does not use any prior data.
    \item \textbf{Init}: This baseline demonstrates the importance of our adaptation procedure during training. Instead of re-adapting the reward model each time new feedback is collected, we initialize the reward model with the pretrained weights, and then perform standard updates with the Adam optimizer \citep{DBLP:journals/corr/KingmaB14} as in PEBBLE. 
\end{enumerate}
For each environment, we reduce the total feedback budget by a factor of 20 in comparison to the maximum value used in PEBBLE. Full results are shown in Figure \ref{fig:metaworld_results}. Overall, we find that despite the 20$\times$ reduction in feedback our method is able to solve almost all of the tasks with a near 100\% success rate. In the Appendix, we directly compare to \citet{lee2021pebble} with using their amount of feedback. In all tasks, except Button Press, we achieve the same asymptotic performance as SAC with 20$\times$ less feedback than originally used for PEBBLE in \citet{lee2021pebble} which is unable to learn a meaningful policy under a reduced feedback budget. In Appendix A we directly compare to PEBBLE with the feedback schedules from \citet{lee2021pebble}. While the \textit{Init} baseline generally performs better than PEBBLE, it still falls short of our method, indicating that re-adaptation is important. Unlike in other pretraining and finetuning paradigms, preference learning is done online, causing the optimal reward function induced by the data to shift. Re-adapting weights each time feedback is collected ensures that we get the full benefits of MAML by considering all data points. Locomotion experiments and ablations on feedback and query selection are included in the Appendix.

\subsection{Can few-shot preference learning be used with humans?}
\label{sec:human}
While no human could be sensibly be expected to provide thousands of pieces of feedback, around a hundred or less not too daunting a task. Given the lower query-complexity of few-shot preference-based RL, we use it to learn complex robot manipulation policies from real-human feedback for the first time. In the process of doing so, we encountered a few challenges. First, humans often have a difficult time answering queries asked by preference based RL algorithms. Queries sampled by maximizing disagreement across an ensemble of reward functions often look identical to humans. Such queries at the margin may be maximally informative, but are more difficult to answer (See Figure \ref{fig:query}). For example, it is unlikely that humans can accurately compare two behavior segments that only have slight variations in the robot joint positions. While this is not explicitly examined in prior work that largely uses artificially generated queries, it is important when considering the abilities of humans and our desired to adapt reward functions with a handful of data points, making everything more sensitive to errors.

To address this, we add the ability for human users to ``skip'' difficult queries instead of providing noisy answers and increased the number of uniform queries used to reduce the likelihood that difficult queries were presented. Second, despite these mitigations, humans still make mistakes in labeling resulting in query labels that are possibly inconsistent. We thus allowed policies to train for longer periods of time between feedback sessions in hopes of collecting more data on the current policy's belief over the reward. \looseness=-1

\begin{figure}[t]
    \centering
    \includegraphics[width=\textwidth]{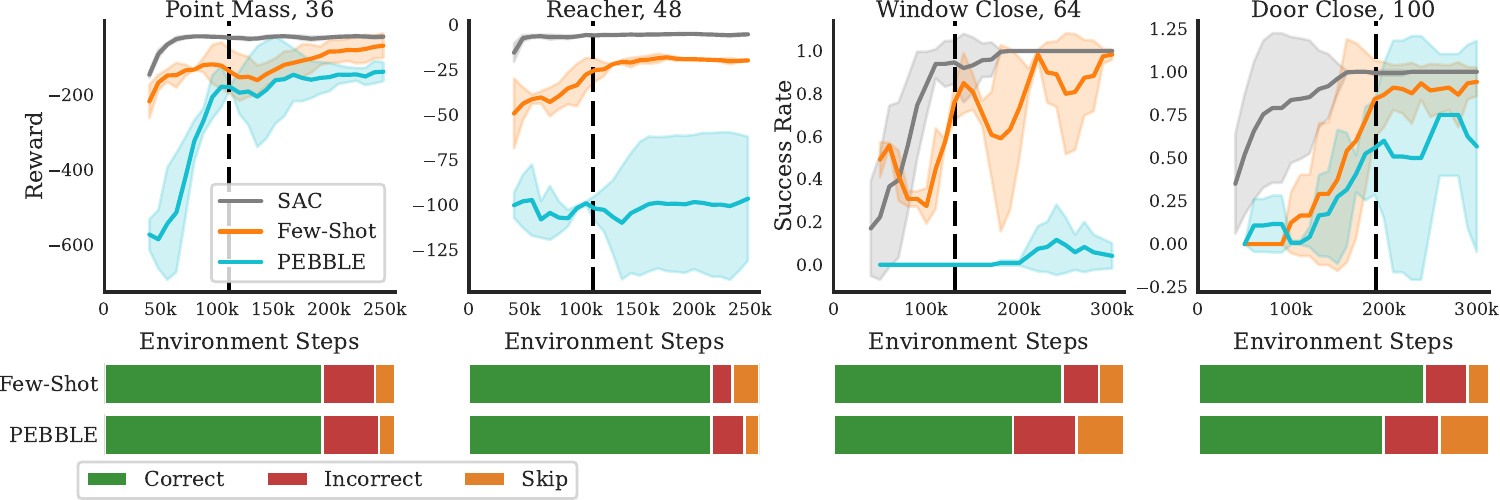}
    \vspace{-0.1in}
    \caption{Results on training DM Control and Meta-World tasks from real human feedback. The vertical dashed black line indicates the point at which feedback was stopped. The horizontal bars at the bottom show the proportion of times users provided feedback that ``correctly'' agreed with the tasks ground-truth reward function, ``incorrectly'' disagree with the ground-truth reward function, or skipped the comparison.}
    \label{fig:human_results}
    \vspace{-0.1in}
\end{figure}

After making the aforementioned changes, we examined the performance of few-shot preference-based RL on two of the MetaWorld environments and two additional environments based on the DM Control benchmark \citep{tunyasuvunakool2020}. We take the point mass and reacher environments from DM Control \citep{tunyasuvunakool2020} and change the reward function to be the negative L2 distance to an unknown goal. Reward models are pre-trained on random data and evaluated on unknown goal positions. The MetaWorld environments are as described in Section \ref{sec:simulation}. Our full results are shown in Figure \ref{fig:human_results}. As the ground-truth reward value for DM control correspond to the cumulative distance to the goal, the higher reward values of our method indicate that it can better communicate the human's objective with fewer queries. While PEBBLE was completely unable to solve Window-Close from human feedback, it made non-trivial progress on Door Close. This is likely because Door Close can be trivially solved by slamming the robot arm into the door instead of first grasping the handle and then closing the door as is encourage by our reward-function prior. Moreover, we find that in the Meta-World environments, users have an easier time answering queries from our method in comparison to PEBBLE. This is likely because the reward function prior guides agents towards interacting with objects, leading to more distinguishable behaviors. Results with more users on Reacher are in Appendix A. \looseness=-1

\begin{figure}[ht]
\begin{minipage}[b]{0.25\linewidth}
\centering
\includegraphics[width=\textwidth]{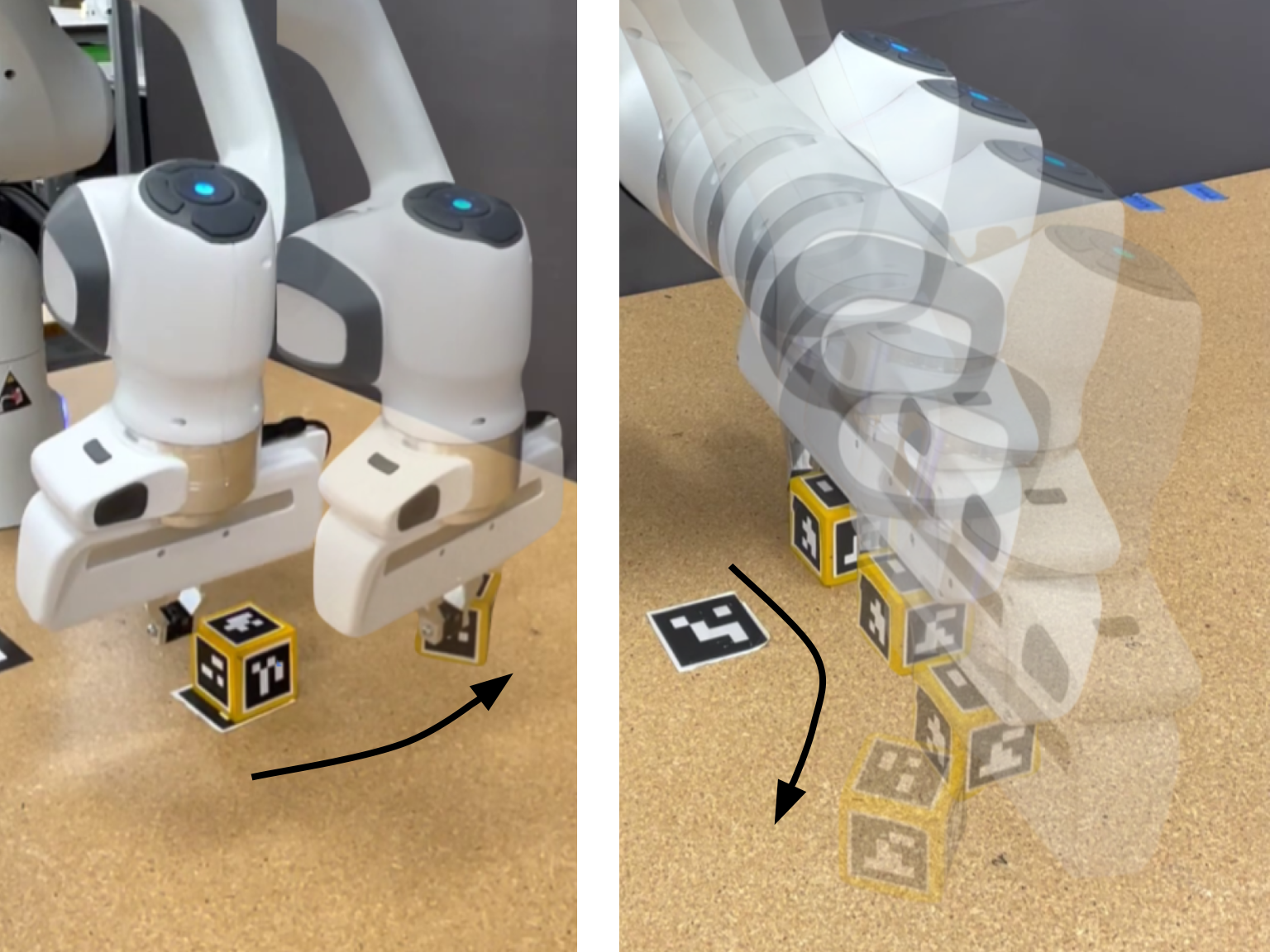}
\vspace{-0.15in}
\caption{Push rollouts}
\label{fig:robot_rollout}
\end{minipage}
\hspace{0.2cm}
\begin{minipage}[b]{0.74\linewidth}
\captionsetup{type=table}
\resizebox{\textwidth}{!}{%
\begin{tabular}{l|llll}
         & \multicolumn{2}{l}{Reach} & \multicolumn{2}{l}{Block Push} \\
Goal Position  & $(.55, .35)$ & $(.45, -.3)$ & $(.35, .3)$ & $(0.35, -0.3)$              \\ \hline
Few-Shot & \textbf{.061 $\pm$ .041} & \textbf{.056 $\pm$ .009} & \textbf{.188 $\pm$ .175} & \textbf{.056 $\pm$ .035}              \\
PEBBLE   & .105 $\pm$ .056 & .129 $\pm$ .067 & .280 $\pm$ .065 & .173 $\pm$ .097
\end{tabular}
}
\vspace{-2pt}
\caption{Results for the real-robot tasks. Performance is measured in meters to the desired goal position, lower is better. The $z$ targets for reach were 0.125 and 0.25, respectively. Results are averaged across multiple initial environment configurations. Best method is bolded.}
\label{tab:robot}

\end{minipage}
\vspace{-0.25in}
\end{figure}

\subsection{Can Few-Shot preference-based RL be used in the real world?}
Finally, we investigate the use of few-shot preference-based RL in real world settings using a Franka Panda Robot. We design two basic tasks: reaching and block pushing where the robot moves its arm or the block, respectively, to an unknown goal location communicated only via the learned reward function. We pre-train reward models with artificial queries and learn policies in simulation. We then transfer the learned policies to the real world and test on unseen goal locations. Table \ref{tab:robot} contains our results. Performance is measured in meters to the true goal. Again, few-shot preference learning consistently outperforms PEBBLE despite the large sim-to-real gap. One additional benefit of our approach in real-world settings is that it potentially requires less instrumentation, as measurements previously needed to functionally compute reward are not required when using human feedback. \looseness=-1

% \section{Limitations}
% discussion of limitations.

% \section{Conclusion}
% A short conclusion

\section{Discussion}

\begin{figure}[t]
    \centering
    \includegraphics[width=\textwidth]{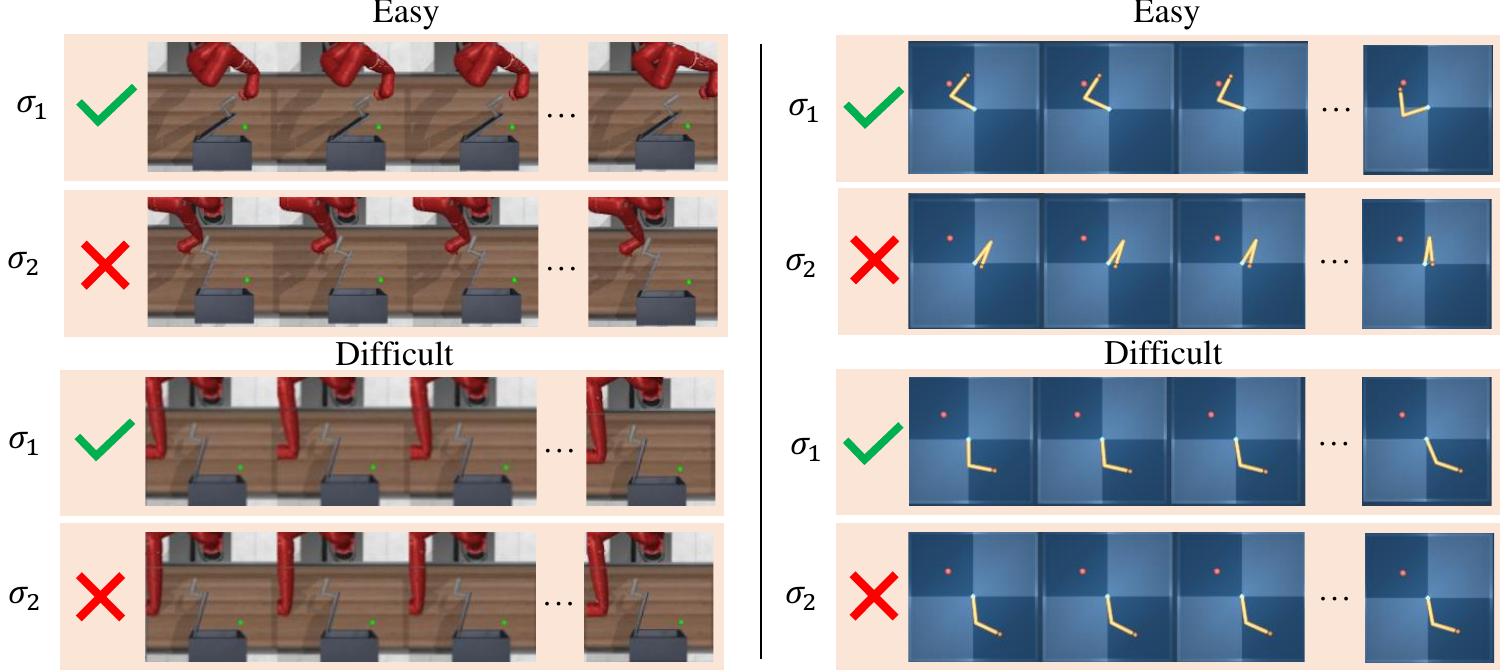}
    \vspace{-0.1in}
    \caption{Examples of queries asked on the MetaWorld Door-Close and DM Control Reacher environments. In each figure the top segment, $\sigma_1$ had higher cumulative reward.}
    \label{fig:query}
    \vspace{-0.1in}
\end{figure}

\textbf{Limitations and Future Work.} 
While few-shot preference learning has several benefits, it has its limitations. Here we list the most salient ones, and possible means of overcoming them: \looseness=-1

\textit{Query Complexity}. Despite our gains in query-efficiency, the most complex tasks still require more feedback than we would like. Future work could examine how to expand the set of pre-training data.  \looseness=-1

\textit{Pre-training Methods }. We investigate pre-training with artificial query data due to its generality, though our method could be used in combination with other pre-training objectives, like direct reward regression, to further boost performance. 

\textit{Pre-training Data}. While meta-learning methods have proven to be somewhat robust to changing dynamics in the real world \citep{clavera2018learning}, the efficacy of reward adaptation under larger distribution shifts induced by sub-optimal users, new tasks, or sim-to-real transfer remains in question. For example, if a new task is significantly out of distribution, we would expect training a reward function from scratch to perform better than adapting. Furthermore, pre-training can occasionally over-regularize the learned reward model, as exhibited in the Door Close experiment in Section \ref{sec:human}.

\textit{Query Difficulty}. Many queries asked by preference learning algorithms are too difficult for humans to answer, as shown in Figure \ref{fig:query}. In fact, we find in Section \ref{sec:human} that active query schemes often result in queries that are too difficult for users to answer. Future work should explicitly consider how easy it is for a human to answer a query and not just its theoretical information content.

\textit{User Inconsistency}. Unlike reward oracles, humans will inconsistently label queries. This challenge is only exacerbated when attempting to crowd source data from many users with differing styles. Future work can investigate additionally modeling human users. \looseness=-1

\textbf{Conclusion.} We shift the paradigm of human-in-the-loop RL from the single-task to the multi-task setting, unlocking additional data sources that can be used to boost the query-efficiency of preference-based RL Algorithms. We believe our work's change in perspective to be a crucial stepping stone towards training robots with human feedback. Our novel few-shot preference-based RL method is able to effectively minimize the number of human queries required to train complex manipulation policies as demonstrated by our 20X improvement on standard benchmarks and effectiveness at real-human training.

%===============================================================================

\clearpage
% The acknowledgments are automatically included only in the final and preprint versions of the paper.
\acknowledgments{This research was supported by NSF (1849952, 1941722, 2218760), ONR, and Ford. JH was supported by the Department of Defense (DoD) through the National Defense Science and Engineering Graduate (NDSEG) Fellowship Program.}

%===============================================================================

\bibliography{references}  % .bib

\newpage
\appendix
\section{Additional Results}

\subsection{Ablations}
\begin{figure}[h]
    \centering
    \includegraphics[width=\textwidth]{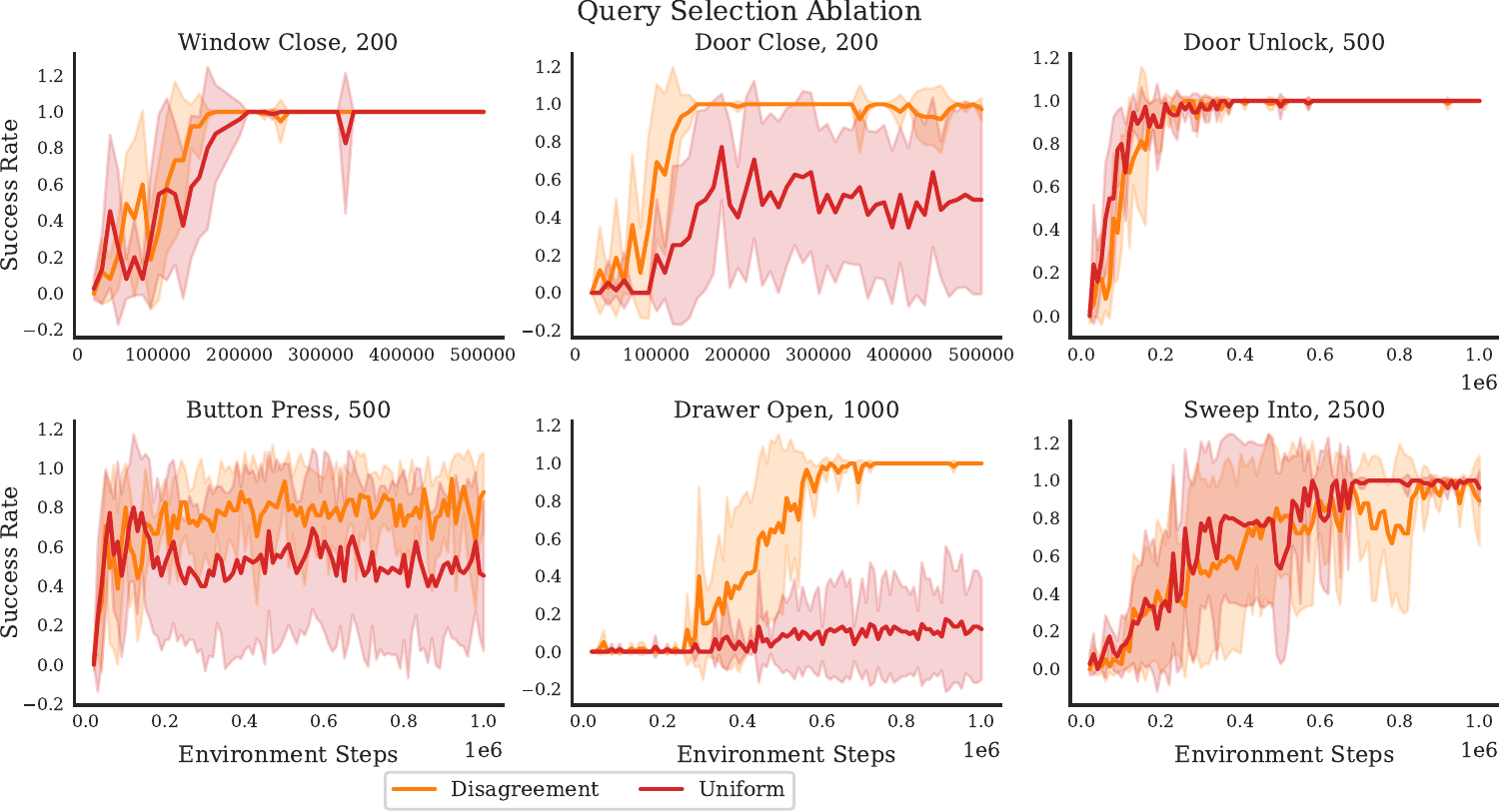}
    \caption{Here we compare using the disagreement query sampling technique versus uniform random query sampling in the MetaWorld environments. We see that for some environments, disagreement sampling is important, but for others it does not have a large effect.}
    \label{fig:disagreement_ablation}
    \vspace{-0.1in}
\end{figure}

In this section we provide a number of additional ablations on the parameters of our method in the MetaWorld environments. Specifically, we vary the amount of total feedback available for both our method and PEBBLE. We train models with PEBBLE using the original amount of feedback in \citet{lee2021pebble}, or 20$\times$ the amount of feedback used in Section \ref{sec:simulation} and Figure \ref{fig:metaworld_results}. Even with 20$\times$ less feedback, our method is at par with PEBBLE. We also train models with our method using only half of the feedback used in Figure \ref{fig:metaworld_results}, and attain nearly the same performance in Window Close, Door Unlock, and Sweep-Into. This indicates that with better parameter tuning, our method could be even more query efficient. Next, we investigate the effects of the disagreement query selection scheme in Figure \ref{fig:disagreement_ablation}. Disagreement sampling leads to performance improvements in some environments, particularly in Drawer Open, but makes no difference in others.

\begin{figure}[h]
    \centering
    \includegraphics[width=\textwidth]{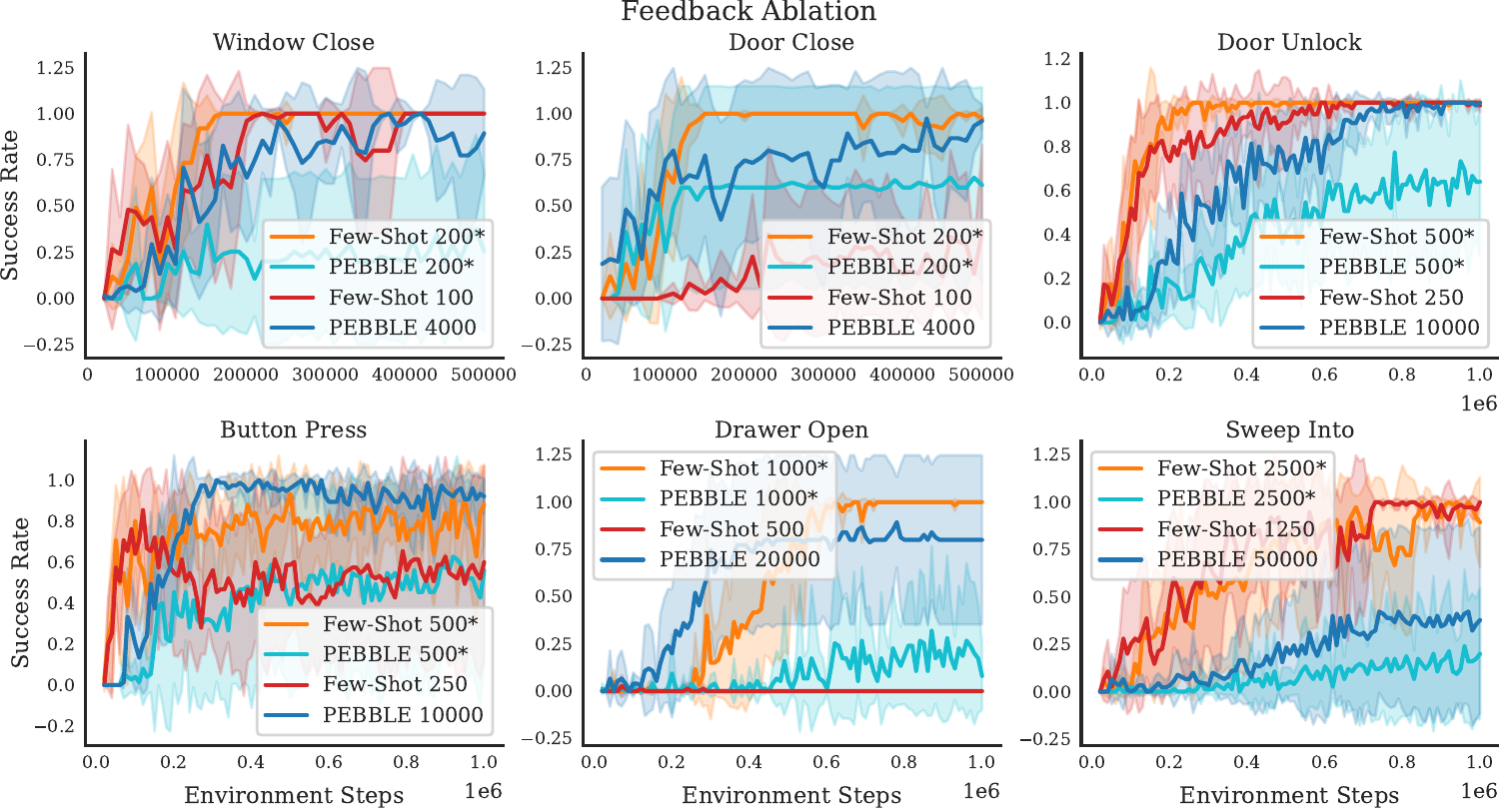}
    \caption{In this ablation, we very the amount of feedback used during training, as indicated by the number in the legend next to each method's name. The ``*'' indicates the original amount of feedback used in Figure \ref{fig:metaworld_results}. We display results using the same amount of feedback as in \cite{lee2021pebble} for PEBBLE, and using half the amount of feedback for our method. Here we can clearly see that our few-shot method performs better than PEBBLE, even though it uses 20$\times$ less feedback. In many tasks, we can half the amount of feedback given to our few-shot method, and still attain the same performance at convergence.}
    \label{fig:feedback_ablation}
    \vspace{-0.1in}
\end{figure}

\subsection{Plots of Feedback versus Performance}
We originally chose to display environment steps on the X-axis of Figures \ref{fig:metaworld_results} and \ref{fig:human_results} as was done in prior work \citep{lee2021pebble, christiano2017deep}. Plotting the environment steps shows the ultimate convergence behavior of each method, as feedback is stopped before the end of training. It also allows us to show SAC on the same graph. Here, we provide versions of Figures \ref{fig:metaworld_results} and \ref{fig:human_results} that have the amount of total feedback given on the X-axis. These plots display the same overall trends -- our few-shot method out-performs baselines for the amount of feedback provided.

\begin{figure}[h]
    \centering
    \includegraphics[width=\textwidth]{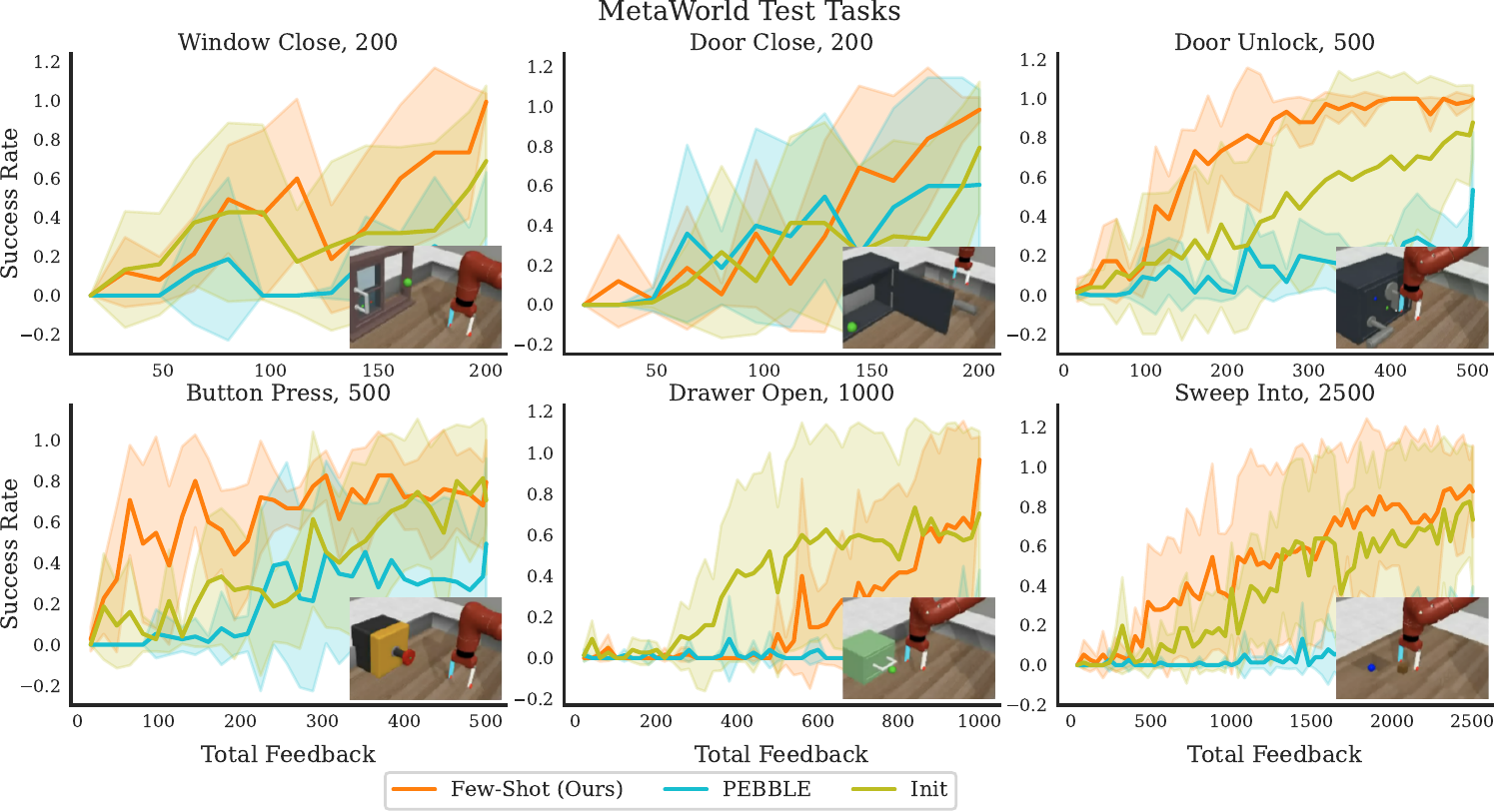}
    \caption{Learning curves for the MetaWorld environments where the x-axis is chosen to be the total feedback given to the agent over the course of training. Note that policies were trained for a bit after all feedback was given, and thus final convergence is not demonstrated as well in this figure, as in Figure \ref{fig:metaworld_results}. In environments where policies obtained decent performance before all feedback was given we were able to further reduce the amount of feedback in the ablation shown in Figure \ref{fig:feedback_ablation}.}
    \label{fig:metaworld_x_feedback}
    \vspace{-0.1in}
\end{figure}

\begin{figure}[h]
    \centering
    \includegraphics[width=\textwidth]{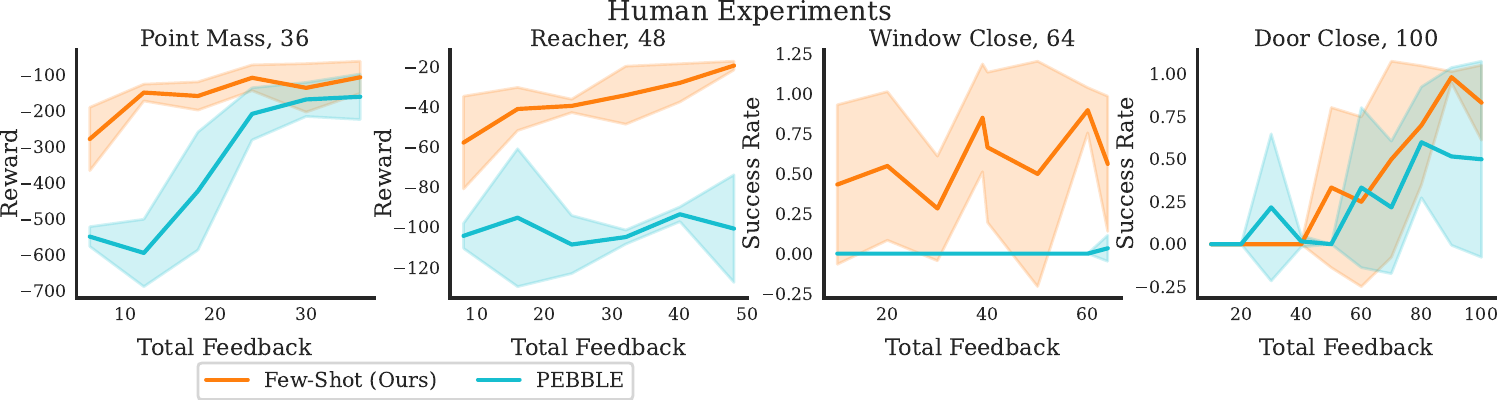}
    \caption{Learning curves for the human user experiments where the x-axis is chosen to be the total feedback given to the agent over the course of training. Again for final convergence, please refer to Figure \ref{fig:human_results}.}
    \label{fig:human_x_feedback}
    \vspace{-0.1in}
\end{figure}

\subsection{Locomotion Experiments}
\begin{figure}[h]
    \centering
    \includegraphics[width=\textwidth]{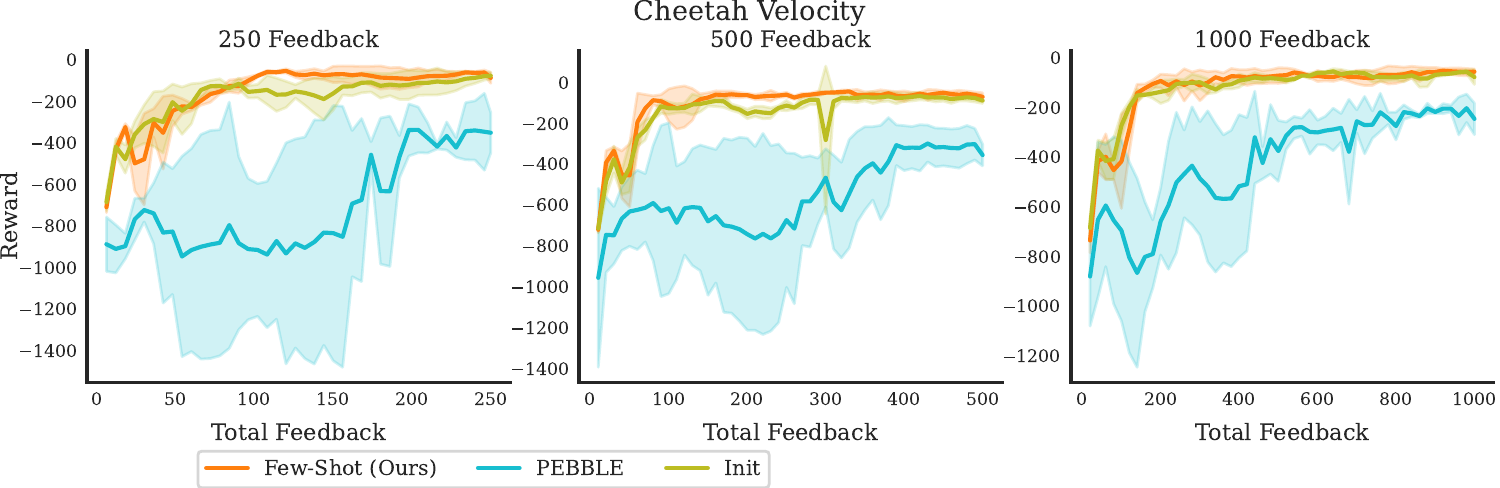}
    \caption{Learning curves for the Cheetah Velocity experiment. The X-axis is given as the amount of feedback provided over 500k environment steps. Each subplot corresponds to a different feedback schedule.}
    \label{fig:cheetah}
    \vspace{-0.1in}
\end{figure}

We evaluate our few-shot preference learning method on a locomotion task, Cheetah Velocity, from \citet{finn2017model} to show its broad applicability, particularly in settings where the agent's goal is temporal and cannot be encapsulated by an environment configuration. The agent is rewarded for moving at a particular unseen target velocity, 1.5m/s. We use 10 other velocities for pretraining. Figure \ref{fig:cheetah} shows our method and PEBBLE using different feedback schedules, with the total feedback provided on the X-axis. Each plot corresponds to training over five-hundred thousand environment steps. We find that our method converges after only around 100 queries independent of the feedback schedule, while PEBBLE is unable to attain close to the same performance even with 1000 queries. The ``init'' baseline described in Section 3 performs similarly, but has slightly worse asymptotic convergence for 2 of 3 feedback schedules. We do minimal hyper-parameter tuning in these environments, and believe the performance of our approach could be further improved. Overall, we find that trends from manipulation environments hold, our few-shot method is able to quickly learn the ground truth reward function. 

\subsection{Comparison with Example Based Methods}
One proposed alternative to inferring reward functions via preferences, is inferring them using examples of ``success'' states to learn reward functions \citep{xie2018few} or directly develop new RL algorithms \citep{eysenbach2021replacing}. While such methods have shown success in their chosen domains, they have a number of drawbacks in comparison to preference based methods. First, example based methods often implicitly assume that the underlying reward function for a task is reaching a goal state. While this is amendable to some tasks, it can preclude objectives that cannot easily be classified as satisfying a goal condition. This is particularly evident for tasks that are temporal in nature, like driving, where we might care about intermediate safety and comfort, not just the final destination. For the aforementioned cheetah locomotion task, it might be difficult for humans to provide examples of successful ``running'' states without a pre-existing oracle policy. While we can easily provide a target velocity, it is difficult to provide target joint positions etc. for a different embodiment. Second, example based methods often optimize sparse-like rewards given for satisfying some learned condition, causing optimization difficulties as horizon scales. This is not the case for preference based methods, which provide consistent dense rewards.

\begin{figure}[h]
    \centering
    \includegraphics[width=\textwidth]{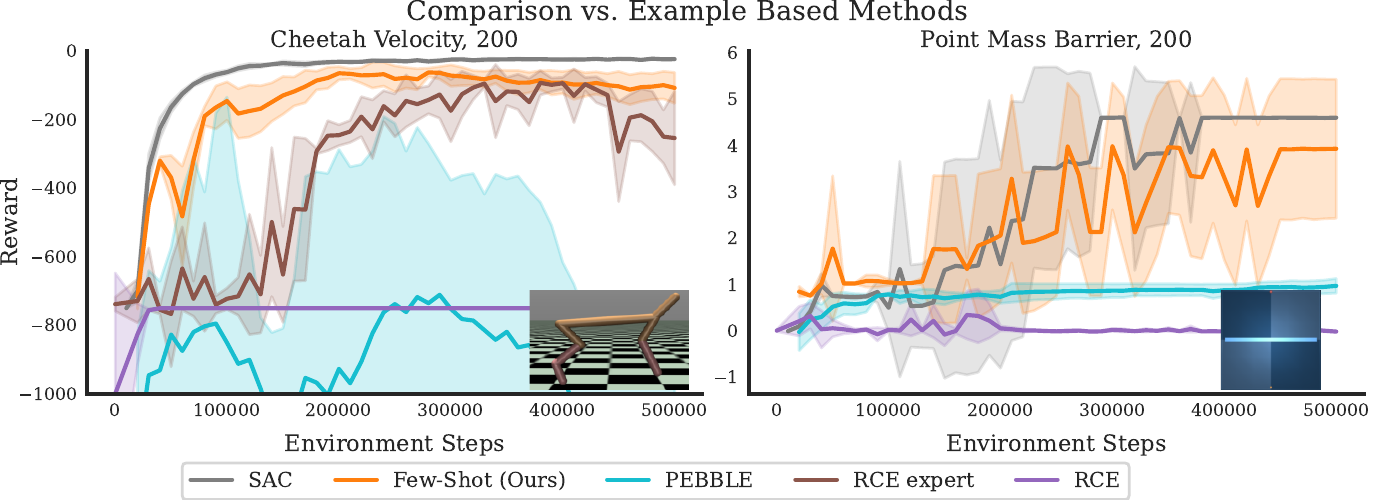}
    \caption{Learning curves for the Cheetah Velocity and Point Maze environment using 200 queries for preference methods and 200 queries for RCE. For the Cheetah environment ``expert'' denotes that examples were generated using a pretrained oracle policy, otherwise examples were generated by relabeling existing data with the target velocity.}
    \label{fig:rce}
    \vspace{-0.1in}
\end{figure}

In order to examine these tradeoffs, we compare our Few-Shot method to Recursive Classification of Examples (RCE) from \citet{eysenbach2021replacing} on two environments using 200 examples or 200 pieces of feedback, though in practice it may be harder to collect examples than preferences. In the Cheetah environment, we examine the effect of example quality on performance by training RCE with states from an expert policy pre-trained with SAC and states from a random policy relabeled to have the target velocity. In a sparse Point Mass Barrier environment, we investigate the impact of horizon and sparsity on example based methods. Results can be found in Figure \ref{fig:rce}. In the Cheetah Velocity environment, we find that even with access to an expert trajectory, RCE does not attain the same asymptotic performance as our method and takes longer to converge. Having access to such data is unrealistic in the real world, as it is impossible to generate success states from a policy if we have not yet solved the task. Even if we had expert demonstrations, it would then perhaps make more sense to directly apply Inverse RL techniques. When we try to train RCE with just states that have been relabeled to the target velocity and do not contain hard-to-specify joint information, performance completely collapses. In the sparse Point Mass Barrier task, we see that despite the 4-dimensional state space RCE is unable to overcome the difficult exploration and long horizon of the task. As our method uses dense rewards learned from preferences, it is almost able to match the oracle SAC policy. While these tasks may be somewhat toy in nature, they demonstrate key areas in which preference based learning excels: when it may be hard to specify temporal behavior via examples, or when tasks are extremely sparse in nature.

\subsection{Human Feedback}
\begin{figure}[h]
    \centering
    \includegraphics[width=\textwidth]{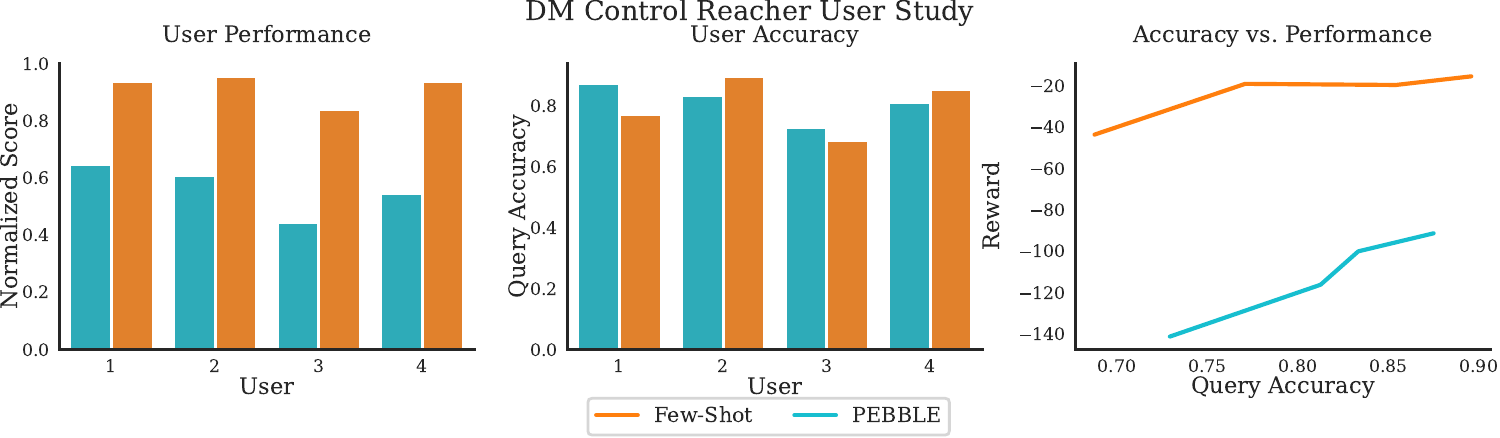}
    \caption{A study of four different users on the DM Control Reacher task. \textbf{Left:} The performance of policies trained by each user expressed as a normalized score between a random policy and a fully trained SAC policy on the task. This is computed as $(\text{method reward} - \text{random reward}) / (\text{SAC reward} - \text{random reward})$. \textbf{Center:} The percentage of each users preferences that aligned with the ground truth reward function for the task. This information was unavailable to the users and is designed to indicate how accurate the human users were. \textbf{Right:} A comparison of final ground truth reward against the alignment of the users preferences with the ground truth reward function.}
    \label{fig:user_study}
    \vspace{-0.1in}
\end{figure}

In order to better understand the effects of different human users on few-shot preference learning, we compare the performance of four different users on the DM Control reacher task. Each user trained one policy using our Few-Shot method and one policy using PEBBLE. The results are shown in Figure \ref{fig:user_study}. Each users provided 48 preferences for each policy. We find that across all users, our few-shot method out performs PEBBLE. Consistent with results in Figure \ref{fig:human_results}, we did not find a significant difference in the difficulty of providing feedback for this task between our method and PEBBLE, unlike in the MetaWorld tasks. Results on the right hand side of Figure \ref{fig:user_study} show that when users preferences do not agree with the ground truth reward function as often, performance declines as expected. Our method is relatively robust until query accuracy, or the amount of time the users preferences agreed with the ground truth reward, dropped below 75\%. At this point, performance began to decline. While these results indicate that our method is robust to human users, it shows a limitation of our work: if users are unable to accurately provide feedback, reward adaptation will suffer.

\subsection{Franka Panda Experiments}
\begin{figure}[h]
    \centering
    \includegraphics[width=0.82\textwidth]{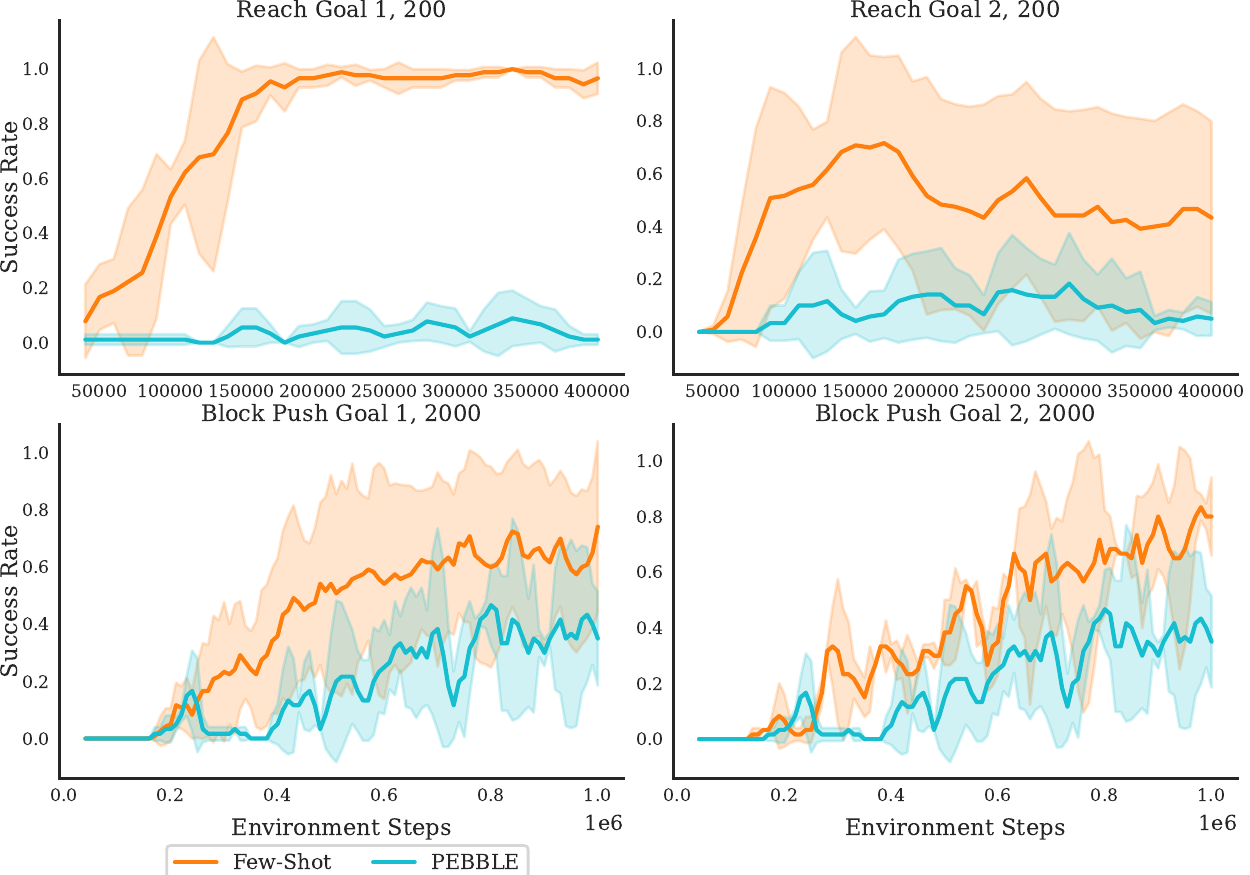}
    \caption{Learning Curves for the Panda experiments in simulation.}
    \label{fig:panda_results}
    \vspace{-0.1in}
\end{figure}

Figure \ref{fig:panda_results} shows the learning curves for the Franka Panda models that could not be fit in the main paper due to space constraints.

\section{Experiment Details}
In this section, we enumerate the specifics of the experiments we use to evaluate few-shot preference based RL. As our method requires generating datasets from past experience, we include dataset generation specifics in addition to environment and evaluation details. 

\subsection{Meta-World}

\textbf{Environments}. For the MetaWorld experiments, we adopt the ML10v2 Benchmark for MetaWorld \citep{yu2019meta}. We keep environments in the ``goal unobserved'' mode, where the agents must infer the final desired position of an object (i.e. door handle) from the reward function alone. MetaWorld environments have both parametric and non-parametric task variations. Parameteric variations refer to changes in the initial and final object positions. Non-parametric variations refer to changes in the objects and their desired conditions, like open door vs close window. Because we wanted to directly compare with the hardest environments used in PEBBLE (Sweep Into, Drawer Open, Button press), we slightly modified the set of environments used in ML10 . This amounts to collecting pre-training data on the 10 tasks shown at the top of Figure \ref{fig:metaworld_results}.

\textbf{Dataset.} The datasets for Metaworld are generated by running ground-truth policies from the 10 prior tasks with some additional Gaussian noise. For each of the 10 tasks, we consider 25 parameteric variations, amounting to 250 different reward functions in the training set, though they each belong to one of only 10 overarching categories. For each of these variations, we collect a dataset of  $(s, a, s', r)$ tuples by running different policies in the given tasks environment with 0 mean, 0.1 standard deviation Gaussian action noise. Specifically, we run 15 episodes with actions from the expert policy, 25 episodes with actions from parametric variations of the same task family, 10 episodes with actions from the expert policy of completely different task family, and 2 episodes with completely uniform random actions. In order to do this we use the scripted policies provided with the MetaWorld benchmark. From each of these datasets, we sample 6000 queries uniformly at random and assign them labels using the ground truth reward. In summary, we use 10 tasks, each with 25 variations of 6000 queries each. 

\textbf{Evaluation.} For MetaWorld, we report the success rate as defined by the MetaWorld benchmark. The test environments are obtained in the same way as PEBBLE, using the standalone versions from MetaWorld. These environments have some parametric variations not included in the prior task environments which makes the test setting slightly more difficult.

\subsection{DM Control} 

\textbf{Environments.} We created custom versions of the standard Point Mass and Reacher environments in DM Control \citep{tunyasuvunakool2020}. The default Point Mass environment has a randomly initialized agent attempt to reach the center of a square environment. We modify the point mass environment so that the goal position is randomly chosen, and use the negative L2 distance to the goal as the ground truth reward function. The default sigmoid style reward function would assign zero reward to a large part of the state space, making artificial query generation difficult. For the reacher environment we mask the goal from the observation space an also use the negative L2 distance as the reward function. All other aspects of the state and action space are left the same.  The point mass environment terminates when the agent reaches the goal position, and the default time limit of the reacher environment was halved to make learning easier. In both of these environments the task distribution is given by the distribution of unknown goal locations. Additionally when comparing to example-based methods we develop a custom Point Mass Barrier environment on top of the standard point mass. We double the size of the point mass environment in both x and y directions, then place a horizontal barrier at $y=0$. The task distribution is also given by different goal locations. The ground truth reward is given by the decrease in L2 distance to the barrier crossing point and then the goal location in sequence (max $< 2$ across the whole trajectory) in addition to a sparse reward of three for reaching the goal. Consequently, the task is considered solved if the agent receives a reward larger than 3. The task distribution is given by goal locations  at $y>0$. 

\textbf{Dataset.} For the Point Mass and Reacher DM control environments we use completely randomly generated dataset. For the Point Mass environment we collect 25,000 random time-steps of the environment 16 X-Y goal positions, which include permutations values in the set $\{0, 0.5, -0.5, 1 ,-1\}^2$. For reacher environment we also collect 25,000 random time-steps of the environment, but over 12 goals each defined by different angle $\theta$ and radius $r$ values, include goals at radius one for each of the four cardinal directions, goals at radius 0.66 for the cardinal directions rotated by 45 degrees, and goals at radius 0.33 for the cardinal directions shifted by 22.5 degrees. From each of the tasks datasets we generate 4000 artificial queries for pre-training uniformly at random. For the Point Mass Barrier task we use 10 pretraining tasks. We then sample 40k queries uniformly at random from the replay buffers of agents train with SAC for 100k steps.

\textbf{Evaluation.} We evaluate the point mass environment on the unseen goal of (-0.75, 0.8) and the reacher environment on the unseen goal of (5.5, 0.8). The Point Mass Barrier task is evaluated on the goal (0, 1) at the top middle of the environment.

\subsection{Franka Panda}
\begin{wrapfigure}{R}{0.5\textwidth}
\centering
    \includegraphics[width=0.49\textwidth]{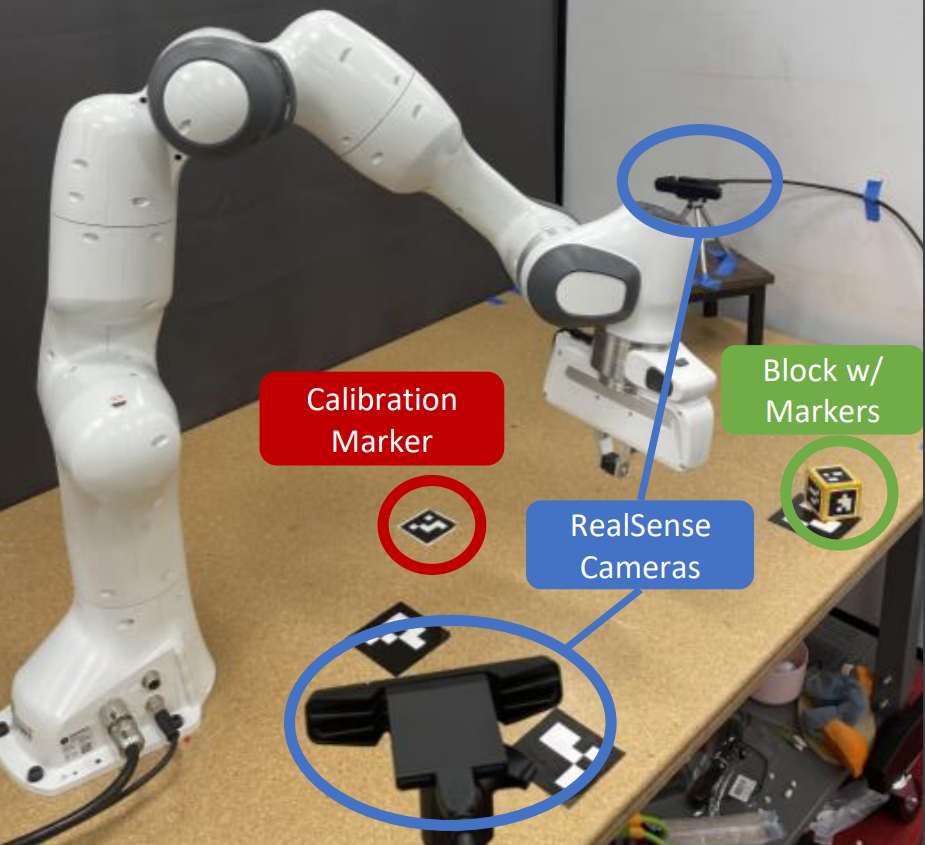}
    \caption{Depiction of the real world robot setup with a Franka Panda arm. We use ArUco tags for tracking the position of objects in combination with Intel RealSense cameras. For the reach task, the robot just needs to move its end effect to a target position. For the block push task, the marked block must be moved to a specific location. The blocks position is computed using two Intel RealSense Cameras.}
    \label{fig:robot_setup}
\end{wrapfigure}

\textbf{Environments.} We design two tasks for the Franka robot. For both tasks we use end-effector delta control, ie the agent chooses x, y, z deltas for the end effector to move to. The first task is the Reach task, where the robot is tasked with simply moving its end-effector towards a target goal position $g$. The reward function is again the negative L2 distance to the goal position, or $- ||e - g||_2^2$ where $e$ is the absolute position of the end effector. The second task is a block pushing task where the agent wants to push a block from a randomized starting location to a fixed goal position $g$. The reward function for this task is $- 0.1||e - b||_2^2 - ||b - g||_2^2$ where $e$ is defined as before and $b$ is the absolute position of the center of the block. The goal positions always have a $z$ value of half the block's height. The agent observes the $(x, y)$ position of the block, but does not know the goal location.  The block is 5cm across. Again the task distribution for both environments is given by the distribution of unknown goal locations. We use the PyBullet simulator for our training environments. When transferring the policies to the real world, use two Intel Realsense cameras and OpenCV Aruco tag tracking to compute the estimated $(x, y)$ position of the center of the block. An image of our setup can be found in Figure \ref{fig:robot_setup}. We also add zero mean, 0.001 standard deviation noise to the state to aid in sim to real transfer. For the Reach task we define success as being within 2.5cm of the goal and for the Block push task we define it to be within 5 cm. 

\textbf{Dataset.} We generate behavior datasets for the reach task by simply collecting random rollouts of 10,000 timesteps for 75 randomly sampled goals. We generate behavior datasets for the block push task by training polices to push blocks to 16 different locations, then applying a similar strategy to the MetaWorld environments: for each task we run 8 random episodes, 50 expert episodes, and 5 episodes using actions from each of the other tasks (80 total), all with zero mean standard deviation 0.3 Gaussian noise. Unlike in meta-world, we did not spend time tuning data generation for the Panda experiments. We then generated 6000 artificial queries for each of the 75 reach tasks, and 20,000 artificial queries for each of the 16 block pushing tasks, leaving one out for validation. 

\textbf{Evaluation.} We evaluate each of the policies by transferring them from simulation to a real Franka-Panda robot. For control, we use the PolyMetis library \citep{Polymetis2021}. We train policies on two different unseen goal positions, which are listed in Table \ref{tab:robot}. We evaluate each run of the reach task using four initial robot configurations and each run of the Block Push task using four initial block locations. Results are reported in final meters to the goal. We found that the second block push location of (0.35, -0.3) was much easier for the robot regardless of method. This is likely because block state estimation was more accurate on that side of the table due to the camera setup.

\subsection{Locomotion}
\textbf{Environments.} We take the Cheetah Velocity environment from \citet{finn2017model}, but use a horizon of length 500. The ground truth reward function is given by $- |v - \text{target}| - ||a||_2^2$, where ``target'' is a target velocity. Thus, the agent is rewarded for running at a certain speed, and we vary the target speed across tasks. Unlike in manipulation environments, reward functions for locomotion environments cannot be specified through any type of ``goal condition'' as behavior across time matters.

\textbf{Dataset.} We generate behavior datasets by taking the replay buffers of policies trained with SAC for 150k environment steps using different target velocities in increments of 0.25m/s, starting with 0.25m/s and ending with 2.75 m/s. We leave out 1.5m/s for the test , making for 10 total training tasks, which is far less than the upwards of 100 training tasks used in \citet{finn2017model}. We generate 40k artificial queries  uniformly at random from each replay buffer for the training dataset.

\textbf{Evaluation.} We evaluate all approaches on the unseen velocity of 1.5m/s.

\subsection{Human Experiments}
Here we provide an overview of the procedure used in our human experiments in Section \ref{sec:human}. We use a single expert human subject for experiments in Figure \ref{fig:human_results}, who was familiar with preference based RL and both the MetaWorld and DM Control benchmarks. The user completed experiments on PointMass, Reacher, Window Close, and Door Close in that order. The human results in Figure \ref{fig:user_study} are from three additional users familiar with learning for robotics, who followed the same procedure. Each environment required training four policies -- two for PEBBLE and two for our Few-Shot method. The user trained all four policies in parallel on a single computer with a user interface that looked similar to the query visualizations shown in Figure \ref{fig:query}. As feedback was elicited intermittently through the course of training, we cannot fully separate the time it took for users to answer queries with the time used to train the policy. However, we know that the total time before all queries were answered was around 22 minutes for Point Mass, around 28 minutes for Reacher, around 45 minutes for Window Close, and around 1 hour for Door Close. Whenever the user could not make a determination about the query, they were asked to skip it. We count skip queries in the total feedback budget and measured the practicality of the user interactions by the number of such skip queries as shown in Figure \ref{fig:human_results}. There we see that human users did not need to skip queries that frequently, and were able to be relatively accurate with respect to the ground truth reward function. Moreover, we found that in the more difficult environments, the human user skipped fewer queries and was more accurate when training a policy using our few-shot method. This is backed up by the visualizations in Appendix D, which qualitatively demonstrates that the few-shot method asks easier to distinguish queries in the robotics environments, likely due to pretraining.

\section{Hyperparameters}
In this section, we detail the hyper-parameters used for our method and baselines. We first give hyperparameters used in pre-training, then provide the hyperparameters used for online experiments. In the following tables we use MW for MetaWorld, DM for DM Control, and FP for Franka Panda. For MetaWorld artificial feedback experiments, we run five random seeds for each method. For human feedback experiments we run two seeds for each method, as it takes a large amount of time to collect human feedback. For real world experiments, we run four seeds for each reaching task, and two seeds for each block pushing task for 8 and 4 seeds total, respectively.

\textbf{Pretraining.} We use the MAML algorithm in combination with the Adam Optimizer. We used learned inner learning rates as in \citet{antoniou2018how}. 

\begin{table}[h]
\centering
\caption{Hyperparameters used for pre-training with the MAML Algorithm.}
\begin{tabular}{cc}
\textbf{Parameter}        & \textbf{Value}  \\ \hline
\rule{0pt}{2.15ex}    Outer LR         & 0.0001 \\
Inner LR         & 0.001  \\
Support Set Size & 32     \\
Query Set Size   & 32     \\
Task Batch Size  & 4      \\
Learn Inner LR   & True  \\
Ensemble Size    & 3 \\
Reward Arch   & 3x 256 Dense \\
Activation     & ReLU \\
Output Activation & Tanh \\
Segment Size     & 25 (MW, FP, C), 10 (DM)
\end{tabular}
\label{tab:meta_params}
\end{table}

\textbf{Online Adaptation.} Here we list the hyperparameters and network architectures used for SAC, PEBBLE, and our method in Table \ref{tab:adapt_params}. In comparison to the original PEBBLE algorithm, we change the segment size to 25 and increased the reward frequency. We found that these changes improved performance for PEBBLE as well. We also train reward models until they achieve 95\% accuracy, instead of training them for a fix number of epochs or until they reach 97\% accuracy as done in the PEBBLE codebase. We run a maximum of 40 MAML adaptation steps. If at that point the reward model has not reached 95\% accuracy, we train it again with the Adam Optimizer. For all methods we did not run unsupervised exploration prior to beginning training. While unsupervised exploration leads to improvements in locomotion environments as shown in \citet{lee2021pebble}, we found that it did not offer a large improvement in robotics environments. This is likely because a sufficient portion of the state space can be explored quickly in locomotion environments like Cheetah and Quadruped, but not in MetaWorld, where task are longer horizon and require both reaching and interacting with specific parts of the state space. For all runs we use a constant feedback schedule, ie the same amount of feedback each session. We list the exact feedback specifications in Table \ref{tab:feedback_params}. Feedback schedules used in the ablation experiments in Appendix A were constructed by multiplying the ``Max Feedback'' and ``Feedback per Session'' values by 20 for PEBBLE and 0.5 for our method.

\textbf{RCE.} For our comparisons against RCE in Figure \ref{fig:rce}, we left all parameters at their defaults. Example states for the Cheetah Velocity environment were given via an expert demonstration, or by relabeling random states with the target velocity. Example states for the Point Mass Barrier environment were created by sampling positions within the target location with feasible velocities.

\begin{table}[h]
\centering
\caption{Hyper-parameters for preference learning algorithms.}
\begin{tabular}{lll}
\textbf{Parameter}                      & \textbf{Artificial Feedback}              & \textbf{Human Feedback}                                 \\ \hline
\rule{0pt}{2.15ex}Init Temp                      & 0.1                              & 0.1                                             \\
Discount                       & 0.99                             & 0.99                                            \\
EMA $\tau$                     & 0.995                            & 0.995                                           \\
Learning Rate                  & 0.0003                           & 0.0003                                          \\
Target Update Freq             & 2                                & 2                                               \\
$(\beta_1, \beta_2)$           & 0.9, 0.999                       & 0.9, 0.999                                      \\
Actor and Critic Arch          & 3x 256 Dense MW, FP, C              & 2x 256 DM,  3x 256 Dense MW                     \\
Actor and Critic Activation & ReLU & ReLU \\
SAC Batch Size                 & 512                              & 512                                             \\
Reward Net Batch Size          & 256                              & 256                                             \\
Disagreement Sample Multiplier & 10                               & 10                                             
\end{tabular}
\label{tab:adapt_params}
\end{table}

\begin{table}[h]
\centering
\caption{Specific feedback schedule for each environment. For all environments, the first session always sampled queries at uniform. For the MetaWorld human experiments, the first half of all queries were asked uniformly at random.}

\begin{tabular}{llll}
\textbf{Environment(s)}          & \textbf{Max Feedback} & \textbf{Feedback Per Session} &\textbf{Session Frequency ($K$)} \\\hline
\rule{0pt}{2.15ex}Window Close, Door Close  & 200          & 8   & 5000                 \\
Door Unlock, Button Press & 500          & 8               & 5000     \\ 
Drawer Open               & 1000         & 10           & 5000    \\ 
Sweep Into                & 2500         & 20           & 5000        \\ 
Point Mass (Human)        & 36           & 6            & 20000        \\ 
Reacher (Human)           & 48           & 8            & 20000        \\ 
Window Close (Human)      & 64           & 8            & 10000        \\ 
Door Close (Human)        & 100          & 10           & 10000        \\ 
Reach Panda               & 200          & 8            & 5000        \\ 
Block Push Panda          & 2000         & 20           & 5000       \\
Cheetah Velocity (vs. RCE)& 200          & 4            & 6000        \\ 
Cheetah Velocity          & 250          & 3            & 5000        \\ 
Cheetah Velocity          & 500          & 5            & 5000        \\ 
Cheetah Velocity          & 1000         & 10           & 5000       \\    
Point Mass Barrier        & 200          & 5            & 10000      \\
\end{tabular}
\label{tab:feedback_params}
\end{table}

\section{Additional Visualizations}
Here we provide select queries shown to users when training from real human feedback using our Few-Shot method. We compare queries asked by each method at the same point in training. The set of nearly all queries used to train agents from human feedback is included in the supplementary material download on OpenReview. Note that the segment size used in MetaWorld was 25, but we showed users every other frame as the changes between individual frames were minimal. In each figure the trajectory segment with the check mark was selected by the user.

\begin{figure}[H]
    \centering
    \includegraphics[width=0.9\textwidth,page=1]{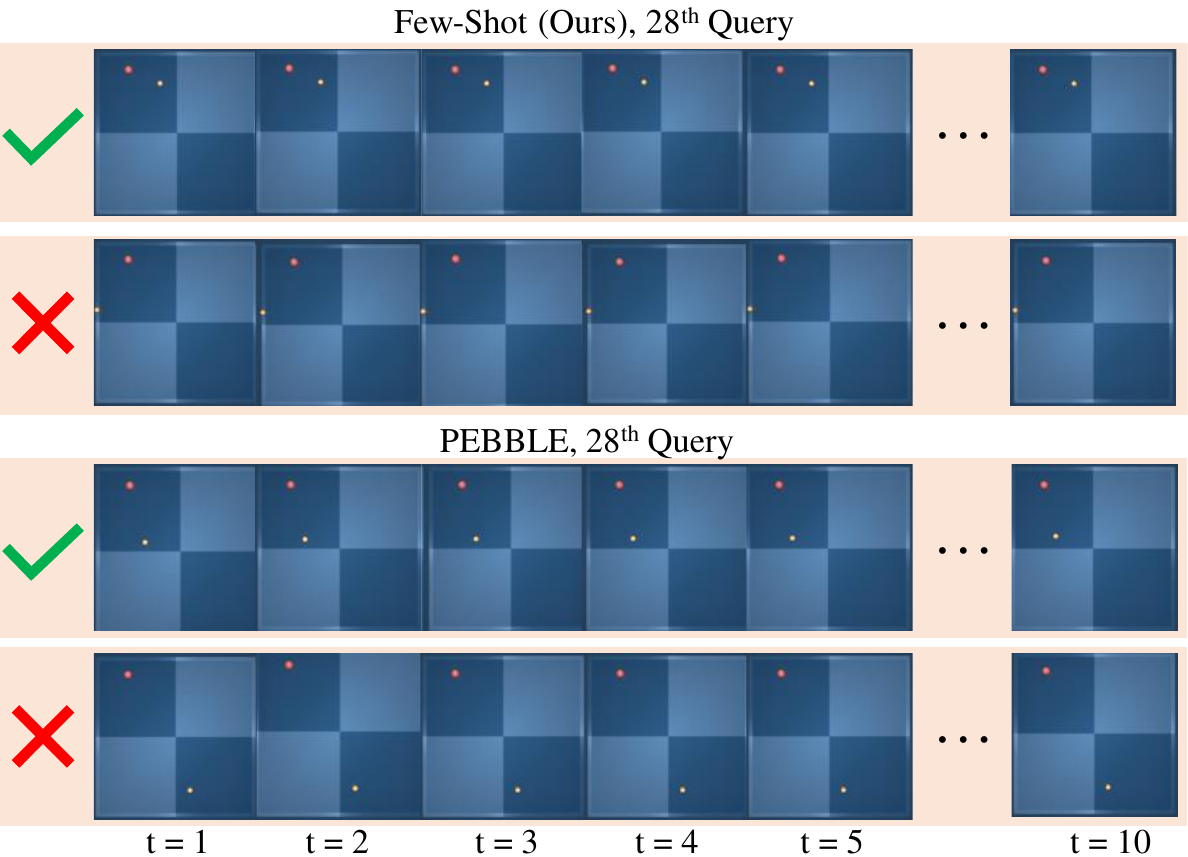}
    \vspace{-0.0in}
    \caption{A depiction of the 28th query asked to users when training the Point Mass Agent from human feedback. The winning query was chosen based on proximity of the agent (yellow) to the goal position (red). At this point in training, our Few-Shot method sampled queries closer to the goal position than PEBBLE.}
    \label{fig:query_pm}
    \vspace{-0.0in}
\end{figure}

\begin{figure}[H]
    \centering
    \includegraphics[width=0.9\textwidth,page=2]{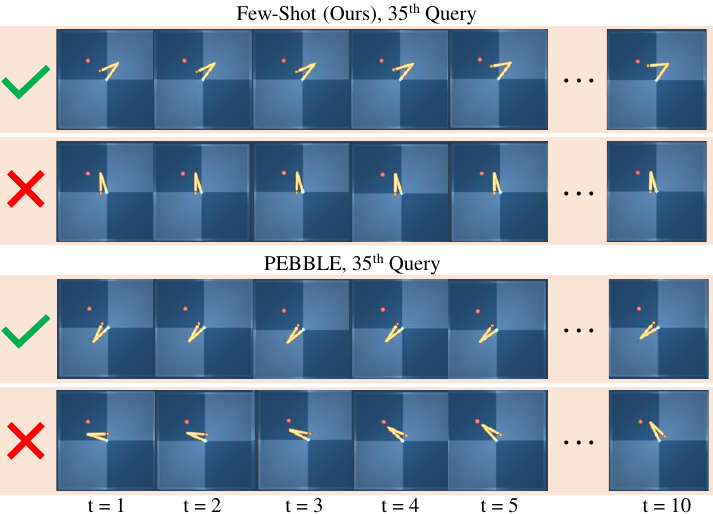}
    \vspace{-0.0in}
    \caption{A depiction of the 35th query asked to users when training the reacher from human feedback. Our method's query (top) was easier to answer because the top trajectories' arm was clearly closer to the target position.}
    \label{fig:query_reacher}
    \vspace{-0.0in}
\end{figure}

\begin{figure}[H]
    \centering
    \includegraphics[width=0.9\textwidth,page=3]{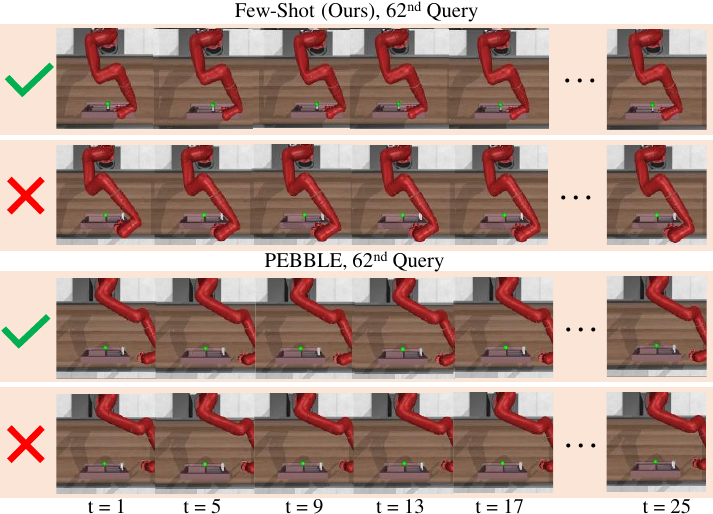}
    \vspace{-0.0in}
    \caption{This shows one of the last queries asked for the Window Close environment. Here we see that our method's query asks the user to choose between a closed and unclosed window (top), while PEBBLE asked the user to choose between two different, hard to distinguish, arm positions.}
    \label{fig:query_window}
    \vspace{-0.0in}
\end{figure}

\begin{figure}[H]
    \centering
    \includegraphics[width=0.9\textwidth,page=4]{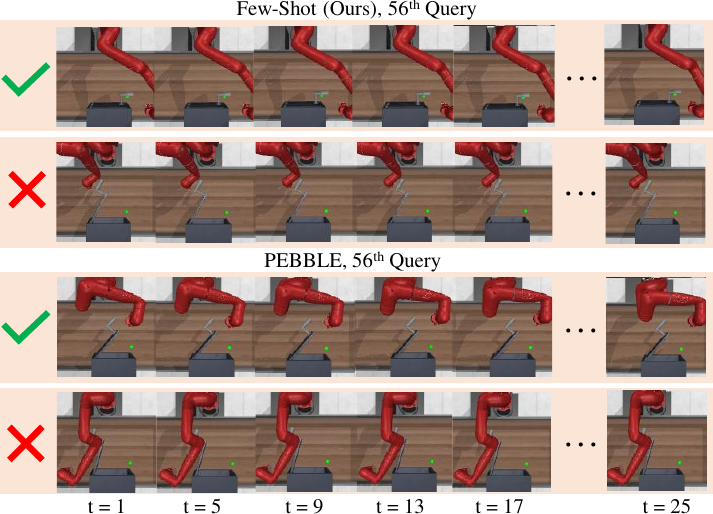}
    \vspace{-0.0in}
    \caption{This shows a query towards the middle of training for the Door Close environment. At this point, the few-shot method is asking the user to compare a completely closed door (better) versus an open one, while PEBBLE's query only includes a partially closed door.}
    \label{fig:query_door}
    \vspace{-0.0in}
\end{figure}

\end{document}